\pdfoutput=1

\documentclass[11pt]{article}
\usepackage{verbatim}
\usepackage[final]{acl}

\usepackage{array}
\usepackage{float}
\usepackage{times}
\usepackage{setspace}
\usepackage{geometry}
\usepackage{latexsym}
\usepackage{algorithm}
\usepackage{algorithmic}
\usepackage[dvipsnames]{xcolor}
\usepackage{subcaption}
\usepackage{makecell}
\usepackage{CJKutf8}
\usepackage{listings}
\usepackage{amsmath}
\usepackage{cleveref}
\usepackage{multirow,booktabs, hhline}
\usepackage{booktabs}
\usepackage{amssymb}
\usepackage{pifont}
\usepackage{bbm}
\usepackage{bbding}

\definecolor{DarkGreen}{RGB}{30,130,30}
\newcommand{\cmark}{\textcolor{DarkGreen}{\ding{51}}}
\newcommand{\xmark}{\textcolor{red}{\ding{55}}}%

\usepackage[T1]{fontenc}
\usepackage[utf8]{inputenc}
\usepackage[nopatch=footnote]{microtype}
\usepackage{inconsolata}
\usepackage{graphicx}
\usepackage{xspace}

\newcommand{\modelname}{\textsc{TinyScientist}\xspace}
\newcommand{\xhdr}[1]{{\noindent\bfseries #1}.} 

\title{\modelname: An Interactive, Extensible, and Controllable\\ Framework for Building Research Agents}

\author{%
  Haofei Yu\textsuperscript{1}\thanks{Core Contributors.} \  
  Keyang Xuan\textsuperscript{1}\footnotemark[1] \  
  Fenghai Li\textsuperscript{1}\footnotemark[1] \  \\
  \textbf{
    Kunlun Zhu\textsuperscript{1}
    Zijie Lei\textsuperscript{1}
    Jiaxun Zhang\textsuperscript{1}
    Ziheng Qi\textsuperscript{1}
   } \\ 
   \textbf{
    Kyle Richardson\textsuperscript{2}
    Jiaxuan You\textsuperscript{1}
  } \\
  \textsuperscript{1}University of Illinois Urbana-Champaign,\\ 
  \textsuperscript{2}Allen Institute for Artificial Intelligence
}

\begin{document}
\maketitle
\begin{abstract}
Automatic research with Large Language Models (LLMs) is rapidly gaining importance, driving the development of increasingly complex workflows involving multi-agent systems, planning, tool usage, code execution, and human-agent interaction to accelerate research processes. However, as more researchers and developers begin to use and build upon these tools and platforms, the complexity and difficulty of extending and maintaining such agentic workflows have become a significant challenge, particularly as algorithms and architectures continue to advance. To address this growing complexity, \modelname identifies the essential components of the automatic research workflow and proposes an interactive, extensible, and controllable framework that easily adapts to new tools and supports iterative growth. We provide an open-source codebase\footnote{The codebase for \modelname is available at \url{https://github.com/ulab-uiuc/tiny-scientist}.}, an interactive web demonstration\footnote{The web demonstration for \modelname is hosted at \url{https://app.auto-research.dev}.}, and a PyPI Python package\footnote{The Python package for \modelname is released at \url{https://pypi.org/project/tiny-scientist}.} to make state-of-the-art auto-research pipelines broadly accessible to every researcher and developer.

\end{abstract}

\section{Introduction}
Interest in building research agents with Large Language Models (LLMs) to interact with human researchers and enable automatic scientific discovery has gained considerable attention in recent years~\citep{gottweis2025towards}. Such agentic frameworks have demonstrated impressive capabilities across a wide range of research tasks, including ideation~\citep{si2024can,li2024learning}, scientific coding~\citep{chan2024mle,huang2023mlagentbench}, paper writing~\citep{wang2024autosurvey}, review writing~\citep{jin2024agentreview}, and even end-to-end research pipelines~\citep{jansen2025codescientist, lu2024ai,yamada2025ai, li2024mlr,cheng2025language}. Recent advances in this area leverage methods including multi-agent collaboration~\citep{schmidgall2025agent}, tool using~\citep{skarlinski2024language}, and tree-based search~\citep{yamada2025ai} to augment its performance.

\begin{figure*}
    \centering
    \includegraphics[width=\linewidth]{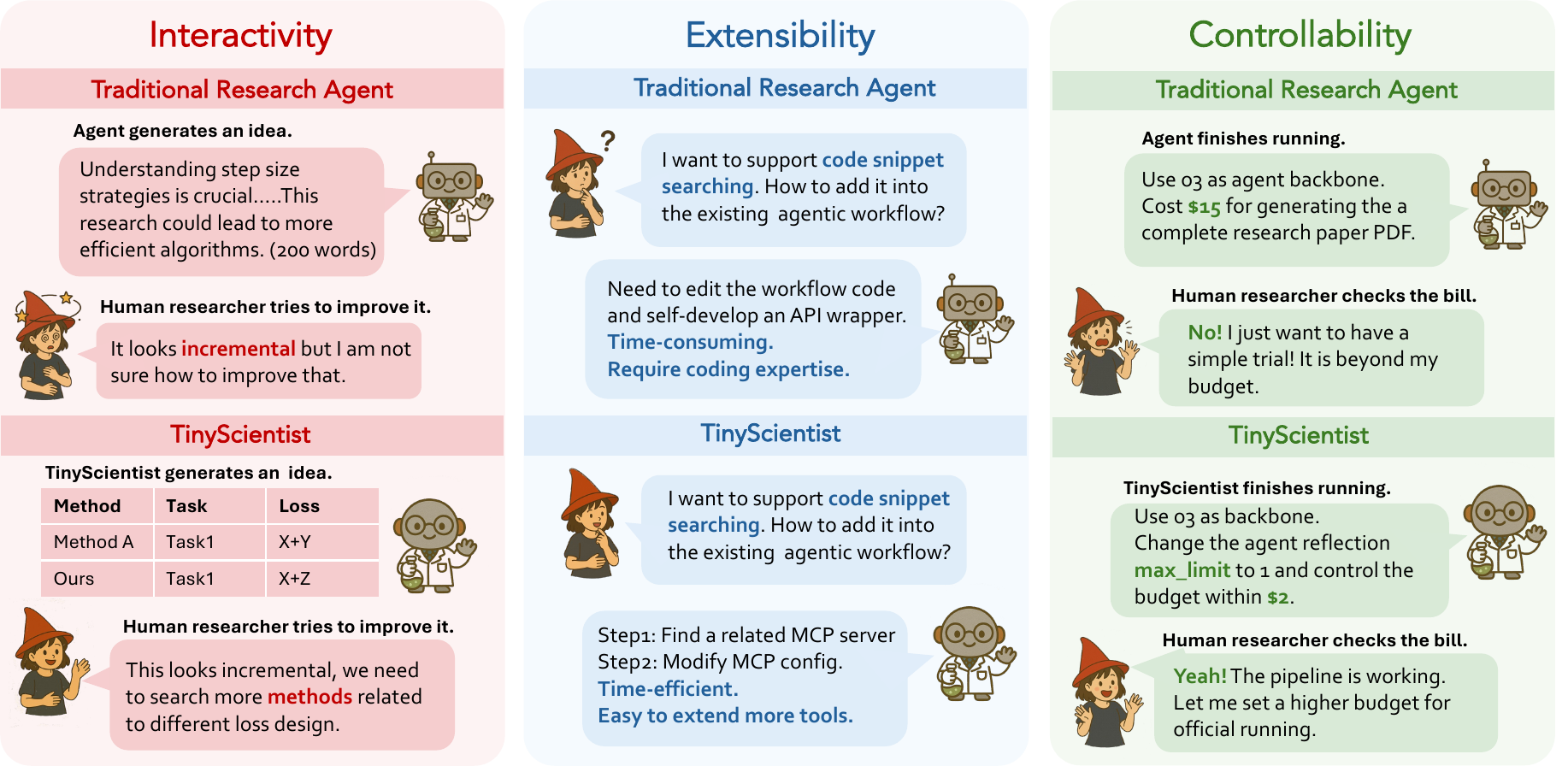}
    \caption{\textbf{Design principles for \modelname.} We highlight the key differences between traditional research agents and \modelname. To enhance \textit{interactivity}, \modelname introduces a table-based interface that helps researchers clearly express and refine their intents. For \textit{extensibility}, \modelname adopts an MCP (Model Context Protocol) design instead of direct API wrapping, making it easy to add or replace tools. For \textit{controllability}, \modelname includes built-in safety and budget controllers that monitor and regulate the entire workflow.}
    \vspace{-1mm}
    \label{fig:design_principle}
\end{figure*}

In spite of this success, however, existing automatic research systems often design and use agentic frameworks that are overly complex and difficult to use and extend without significant technical expertise. These challenges stem from three key issues:
(1) \textbf{\textit{lack of interactivity}}: human researchers struggle to engage with the specific agent's research progress due to the complexity of research intents and unclear communication interfaces \citep{zou2025survey, liu2025real}, making feedback incorporation challenging.
(2) \textbf{\textit{limited extensibility}}: existing representative frameworks rely on rigid, tool-specific designs \citep{zhang2025agentorchestra}, making it hard to integrate new tools or adapt to different research domains.
(3) \textbf{\textit{insufficient controllability}}: many systems offer weak supervision on safety, ethical constraints, and cost budget, raising concerns about misalignment and unbounded execution \citep{gridach2025agentic, liu2025advances}. To address these issues and help democratize the development and use of research agents, we introduce \modelname, a lightweight and modular agentic framework that facilitates interactivity, extensibility, and controllability, making it highly accessible to users, researchers, and developers. Specifically, \modelname is designed based on the following principles illustrated in Figure~\ref{fig:design_principle}.

\vspace{1mm}
\xhdr{Interactivity}
Research is an open-ended process that requires continuous user engagement. Researchers typically begin with vague or evolving goals and refine them over time. As a result, effective research agents must support real-time adjustments to their reasoning, coding, and writing processes. Without an interactive interface, agentic frameworks risk drifting from the user's intent and producing irrelevant or unsafe outputs. To address this, \modelname introduces a modular, tabular-based interface that decomposes the research workflow into editable stages. Each stage presents intermediate results in a structured tabular format, allowing researchers to directly modify specific cells or columns—\textit{e.g.}, by suggesting adding new baselines as rows or editing individual entries. This design improves clarity in human-agent communication and empowers users to guide the system as their objectives evolve.

\vspace{1mm}
\xhdr{Extensibility} 
Automatic research is evolving rapidly, with new tools and technologies emerging constantly. To keep pace, agentic frameworks need to support the easy integration and replacement of tools. In machine learning-related automatic research, while the core workflow—typically composed of stages like think, code, write, and review—remains relatively fixed, the key difference between different tasks lies in the tools and methods used within each stage. To address this, \modelname adopts the design of the Model Context Protocol (MCP)~\citep{anthropic2024mcp}, which provides a unified API for connecting diverse tools to augment each core workflow component. Such a system architecture enables seamless extension, allowing developers to upgrade and maintain their system with little effort.

\begin{table*}[!t]
\setlength\tabcolsep{2pt}
    \centering
    \small
    \vspace{-3mm}
    \resizebox{\linewidth}{!}{
    \begin{tabular}{l|c|c|c|c|c|c|c|c}
        \toprule
        \multirow{2}{*}{\raisebox{0.5\normalbaselineskip}[0pt][0pt]{\textbf{Framework}}}
        & \multicolumn{2}{c|}{\textbf{Interactivity}} 
        & \multicolumn{2}{c|}{\textbf{Extensibility}}
        & \multicolumn{2}{c|}{\textbf{Controllability}} 
        & \multicolumn{2}{c}{\textbf{Deployment}} \\
        \cmidrule(lr){2-3} \cmidrule(lr){4-5} \cmidrule(lr){6-7} \cmidrule(lr){8-9}
        & \makecell{\textbf{Modular} \\ \textbf{Design}} 
        & \makecell{\textbf{Tabular} \\ \textbf{Commun.}} 
        & \makecell{\textbf{Tool} \\ \textbf{Calling}}
        & \makecell{\textbf{Schema} \\ \textbf{Diagram}}
        & \makecell{\textbf{Safety} \\ \textbf{Control}} 
        & \makecell{\textbf{Budget} \\ \textbf{Control}} 
        & \makecell{\textbf{UI} \\ \textbf{Design}}
        & \makecell{\textbf{Python} \\ \textbf{Package}}\\ 
        \midrule
        AI Scientist~\cite{lu2024ai}  & \cmark & \xmark & API & \xmark & \cmark & \xmark & \xmark & \xmark\\ 
        AI co-scientist~\citep{gottweis2025towards}  & \cmark & \xmark  & API & \xmark & \xmark & \xmark & \cmark & \xmark\\ 
        AI Researcher~\citep{tang2025ai}  & \cmark & \xmark & API & \xmark & \xmark & \xmark & \xmark & \xmark\\ 
        Agent Laboratory~\cite{schmidgall2025agent}  & \cmark & \xmark & wrapper & \xmark & \cmark & \xmark & \xmark & \xmark\\ 
        \midrule
        \textbf{TinyScientist (ours)}  & \cmark & \cmark &  MCP & \cmark & \cmark & \cmark & \cmark & \cmark \\ 
        \bottomrule
    \end{tabular}}
    \caption{\textbf{Comparison of research agent frameworks.} We compare existing frameworks across four key dimensions: interactivity, controllability, extensibility, and deployment readiness. \modelname uniquely integrates tabular-based human-agent communication, schematic diagram design, MCP-based tool calling, and budget/safety control, all within a deployment-ready system. For tool calling, \textit{API} refers to frameworks that directly invoke APIs as part of their workflow without abstraction. In contrast, \textit{wrapper} denotes systems that support tool abstraction and allow users to wrap custom tools via APIs, though without a standardized integration method like MCP.}
\label{tab:framework_comparison}
\end{table*}

\vspace{1mm}
\xhdr{Controllability}
Safety, ethical, and financial concerns are critical, but often under-addressed in agentic research workflows~\citep{zhu2025safescientistriskawarescientificdiscoveries,tang2024prioritizingsafeguardingautonomyrisks}. Users should not be caught off guard by excessive spending or unsafe outputs. Users should not be exposed to unexpected costs, unsafe behaviors, or misaligned actions. To ensure responsible and predictable execution, \modelname emphasizes controllability across the entire pipeline. Users are allowed to set explicit upper bounds on key hyperparameters—such as the number of experimental runs or self-reflection steps—to ensure budget constraints are respected. Additionally, at each stage of the workflow, built-in safety checkers validate outputs to prevent harmful or unintended behavior, which helps to maintain alignment with user intents.

To demonstrate the effectiveness of \modelname, we conduct both qualitative and quantitative evaluations, highlighting four key advantages:
(1) It is easy for researchers to use the Python package without configuration or setup barriers.
(2) It enables better human–agent interaction through an interactive UI.
(3) It achieves research generation quality comparable to Agent Laboratory~\citep{schmidgall2025agent}, a widely used multi-agent auto-research framework, and (4) tool usage improves the generation quality.

\section{Related Work}

\xhdr{Agentic workflow for automatic research} 
The field of automatic research has witnessed rapid advancements in recent years, with diverse frameworks emerging to automate the research process through various design principles and coordination strategies. For example, AI-Scientist~\citep{lu2024ai} introduced the first comprehensive fully-automatic research agent by enabling frontier LLMs to conduct a series of research processes. In subsequent work,  AI-Scientist v2~\citep{yamada2025ai} improves the pipeline by replacing manual templates with an agentic tree-search methodology and a VLM-based feedback loop. In addition, later work such as Agent Laboratory~\citep{schmidgall2025agent} and AI-Researcher~\citep{tang2025ai} follow a similar staged design to develop end-to-end autonomous research workflows, introducing more refined role specialization and enhanced coordination mechanisms. Furthermore, AI co-scientist~\citep{gottweis2025towards} and ResearchTown~\citep{yu2024researchtown} utilize a specialized multi-agent coordination paradigm to facilitate novel scientific idea discovery. While prior work pursues automation through complex orchestration, our work prioritizes simplicity and modularity by distilling the research process into four core stages, striking a balance between automation, simplicity, and extensibility.

\vspace{1mm}
\xhdr{Human-in-the-loop for automatic research} While fully automated research pipelines are promising, they remain practically limited without human involvement, underscoring the need for human oversight. Recognizing this, recent work has incorporated human-in-the-loop functionality that allows researchers to contribute to different stages of automated research. For the idea stage, \citet{garikaparthi2025iris} and \citet{radensky2024scideator} support interactive hypothesis refinement and facet recombination with researcher feedback. In addition, CodeScientist~\citep{jansen2025codescientist} introduces an end-to-end system for semi-automated scientific discovery where humans can collaborate with LLMs to design, execute, and interpret code-based scientific experiments. Furthermore, \citet{ifargan2025autonomous} and DeepReview~\citep{zhu2025deepreview} incorporate human experts' feedback to review and refine LLM-generated scientific drafts, ensuring alignment with expert judgment. Our work follows this trend by treating human feedback as the central component, particularly through our table-based user interface design.

\section{\modelname Framework}
In this section, we first provide a brief overview of \modelname, including its input and output and overall architecture design. We then describe each core module in the agentic workflow backbone. Finally, we introduce the features built on top of this workflow that make \modelname interactive, extensible, and controllable.

\begin{figure*}
    \centering
    \includegraphics[width=\linewidth]{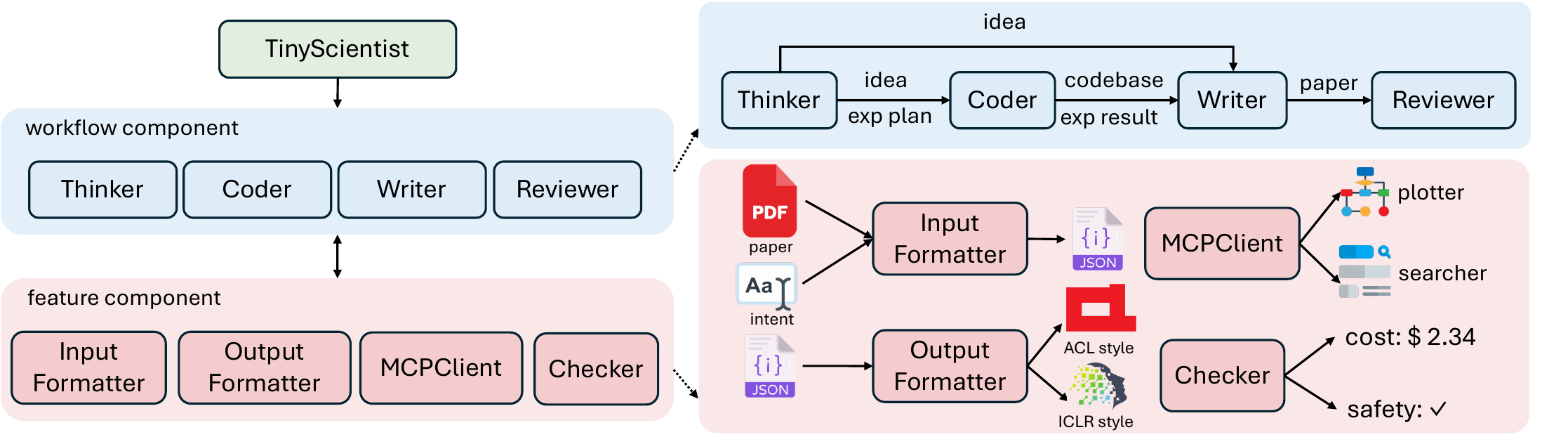}
    \caption{\textbf{Overview of \modelname framework}. On the left side, the diagram illustrates the class hierarchy of \modelname. At the top-left side, \texttt{TinyScientist} serves as the base class. It manages four workflow components, each responsible for a core stage of the research process. In turn, each workflow component is supported by four feature components that enhance its functionality beyond its core function. On the right side, the overall workflow and the details for each feature component are described separately.}
    \label{fig:framework-overview}
\end{figure*}

\subsection{Framework Overview}
The ultimate goal behind \modelname is to minimize the complexity of research agent workflows and make them accessible to everyone. To achieve this, we first clarify the input/output design of our framework. We then describe the hierarchical and modularized component architecture included in our framework.

\vspace{1mm}
\xhdr{Framework I/O} To support general use, \modelname accommodates multiple input and output formats. We identify three common types of research data: AI conference-style PDFs (\textit{e.g.}, ACL\footnote{We refer to using the ACL conference template as \url{https://github.com/acl-org/acl-style-files}} and ICLR\footnote{We refer to using the ICLR conference template as \url{https://github.com/ICLR/Master-Template}}), structured JSON data, and plain text strings. The I/O design of \modelname accepts any of these formats as input and can generate output in any of them. For example, a typical use case is taking a plain-text intent as input and producing a fully formatted conference paper PDF as output.

\vspace{1mm}
\xhdr{Framework architecture}
As shown in Figure~\ref{fig:framework-overview}, our framework is organized into a clear, hierarchical architecture. At the top, the Engine class orchestrates four core \textit{workflow components} (thinker, coder, writer, and reviewer), and each represents a distinct stage in the research lifecycle. These components are modularized for interactivity, allowing users to inspect and guide the agent at each step. Each workflow component is further supported by a set of reusable \textit{feature components}, including InputFormatter, OutputFormatter, MCPClient, and Checker. These components are designed with specific goals in mind: extensibility through MCPClient for flexible tool integration, and controllability through the Checker for enforcing safety and budget constraints. Together, the separation between workflow and feature components ensures a clean framework architecture.

\subsection{Workflow Components}
\label{subsection:workflow-components}
In this section, we first describe the overall agentic structure and the specific functionality of each workflow component.

\vspace{1mm}
\xhdr{Basic: Iterative agent} Each agent follows an iterative, self-refinement paradigm~\citep{renze2024self, madaan2023self}, where it repeatedly performs and improves upon a task until reaching a predefined iteration cap. This shared iterative structure applies to all stages in the workflow.

\vspace{1mm}
\xhdr{Stage1: Think} The \textit{Thinker} module is responsible for research ideation based on the user’s input intent. It samples $n$ initial ideas via LLM-based prompting, where each idea is refined through $k$ rounds of iterative improvement. Each idea contains: (1) a descriptive paragraph; (2) an experimental plan; (3) a comparison table with related works. Additionally, the Thinker provides self-evaluation scores for each idea along three dimensions: impact, feasibility, and novelty. Formally, we express this process as the following transformation:
\begin{equation*}
\texttt{Thinker}(\texttt{intent}) \rightarrow \texttt{idea}
\end{equation*}

\xhdr{Stage2: Code} Given an idea and its experimental plan, the \textit{Coder} module leverages an external coding agent framework (\textit{e.g.}, \texttt{Aider}\footnote{We refer to \url{https://github.com/Aider-AI/aider}}) to iteratively generate executable codebase and run experiments. If the execution fails or deviates from the plan, the agent pauses and awaits human input. Conversely, upon successful execution, it returns the experimental results and associated code artifacts to the output directory. Abstractly, this takes the following form:
\begin{equation*}
\texttt{Coder}(\texttt{idea}) \rightarrow \texttt{codebase}
\end{equation*}

\xhdr{Stage3: Write}
The \textit{Writer} module treats scientific paper writing as a structured three-step process:
(1) initial generation,
(2) paper refinement, and
(3) citation insertion.
For each paper section (\textit{e.g.}, Introduction), the writer receives structured inputs, such as an idea and a codebase, then generates a draft using an LLM. The draft is then refined based on an error checklist to fix \textsc{LaTeX} format inconsistencies. Finally, citation embedding is performed by retrieving relevant references $\{ r_1, r_2, \ldots, r_n \}$ via the Semantic Scholar API \citep{kinney2023semantic}\footnote{We refer to \url{https://api.semanticscholar.org/}}, and inserting them into the completed draft to yield the final version. As above, we can define this as:
\begin{equation*}
\texttt{Writer}(\texttt{idea}, \texttt{codebase}) \rightarrow \texttt{paper}
\end{equation*}

\xhdr{Stage4: Review} The \textit{Reviewer} module evaluates the completed paper and simulates peer reviews, each including a summary, strengths, and weaknesses in standard format. Every review is refined through self-reflection, and a meta-review is synthesized along with a final score:
\begin{equation*}
\texttt{Reviewer}(\texttt{paper}) \rightarrow \texttt{review}
\end{equation*}

\subsection{Feature Components}
\label{subsection:feature-components}

Beyond the four workflow modules discussed in Section~\S\ref{subsection:workflow-components}, \modelname introduces three key feature components—Formatter, MCPClient, and Checker—to support its core principles: interactivity, extensibility, and controllability, respectively.

\begin{figure}[!t]
    \centering
    \includegraphics[width=\linewidth]{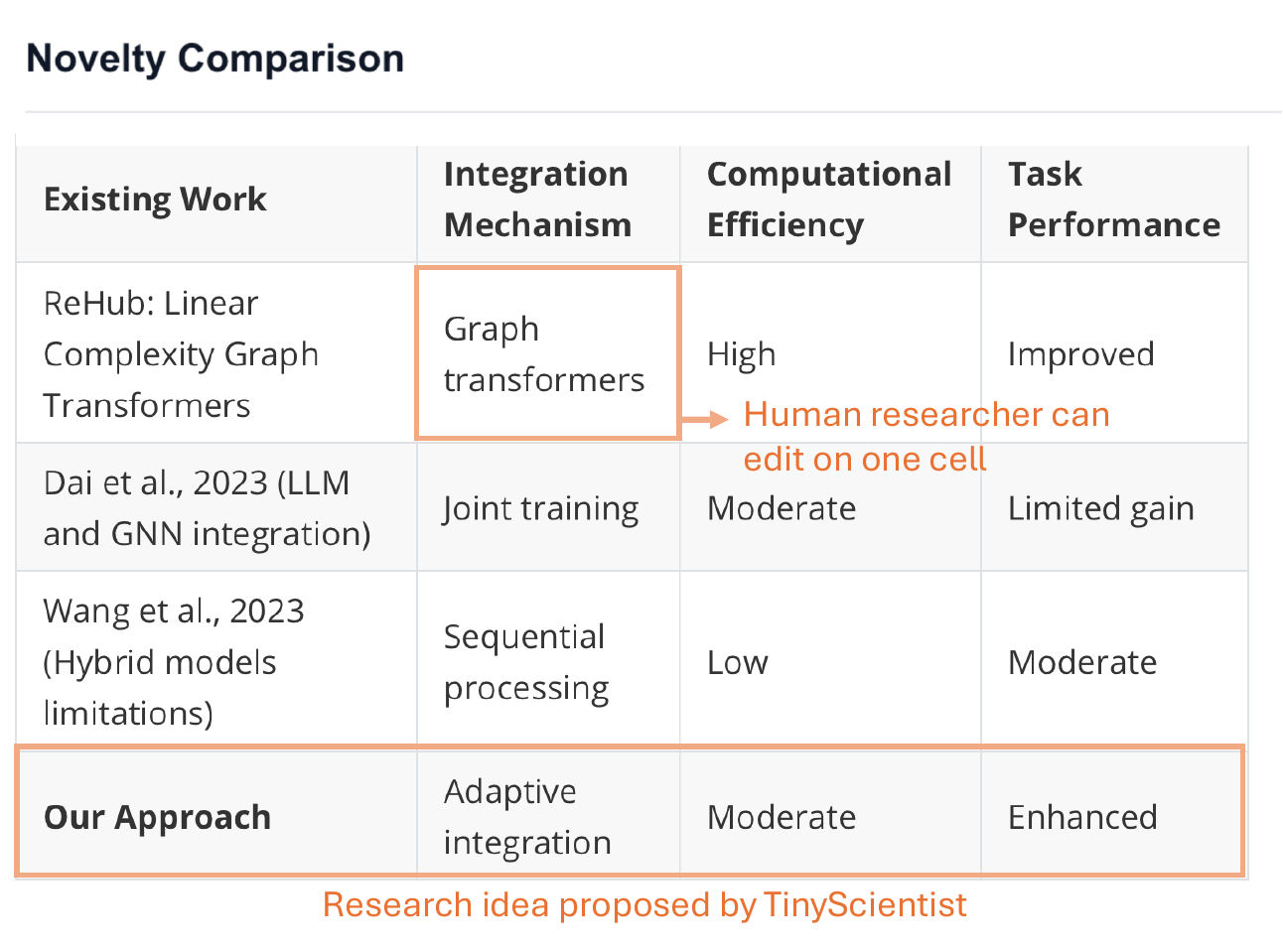}
    \caption{\textbf{Example for tabular-based interaction between stages}. This shows one novelty comparison result of idea thinking. It organized the generated idea as one line within one table.}
    \label{fig:ui-feature2}
    \vspace{-4mm}
\end{figure}

\vspace{1mm}
\xhdr{Formatter: Enhancing interactivity via tabular-based communication}
In automatic research, agents often generate large amounts of intermediate information, making it difficult for human researchers to track progress and monitor individual steps. To address this, \modelname uses the Formatter to compile key outputs into structured tables between different workflow components. These tables provide a clear summary and comparison of the agent’s progress, enabling researchers to easily review, comment on, and directly edit specific elements, thereby facilitating precise and interactive guidance throughout the workflow. Figure~\ref{fig:ui-feature2} shows a concrete example of the table generated by LLMs for research ideation.

\vspace{1mm}
\xhdr{MCPClient: Enhancing extensibility via tool integration}
Modern research workflows require support beyond LLM prompting. MCPClient serves as a bridge between workflow components and a wide range of research tools—such as code searchers, plot drawers, and paper retrievers—enabling seamless integration and future extensibility. Appendix~\S\ref{case-study} provides an example of how MCP is used to enhance paper writing by incorporating schematic diagram generation.

\vspace{1mm}
\xhdr{Checker: Enhancing controllability via budget and safety constraints}
To ensure controllable usage, the Checker module enforces constraints on cost and safety. Users can set limits (\textit{e.g.}, model size, max iterations), and the system adjusts parameters like the number of self-reflections accordingly. Stage-specific safety filters (\textit{Thinker}, \textit{Coder}, \textit{Reviewer}) proactively block harmful outputs. Table~\ref{tab:safechecker_result} shows an example for the safety checker.

\section{\modelname Python Package}
To demonstrate the practicality of \modelname, we develop a Python package based on the proposed agentic framework, enabling easy and modular use. As illustrated in Algorithm~\ref{alg:tinyscientist}, with just seven lines of code, the package can generate a fully compiled PDF of a complete research paper in the standard AI conference format - along with the research idea, the corresponding experimental code, and details of the peer review process.

\begin{algorithm}[ht]
\caption{TinyScientist usage example}
\label{alg:tinyscientist}
\small  
\begin{algorithmic}[1]

\STATE \textcolor{CadetBlue}{\texttt{\# model: string name for LLM}}
\STATE \textcolor{CadetBlue}{\texttt{\# intent: string of user intent description}}
\STATE
\STATE \texttt{\textcolor{red}{from} tiny\_scientist \textcolor{red}{import} \textcolor{Mahogany}{TinyScientist}}
\STATE \texttt{\textcolor{CadetBlue}{\# Instantiate TinyScientist}}
\STATE \texttt{scientist = \textcolor{Mahogany}{TinyScientist}(model)}

\STATE \texttt{\textcolor{CadetBlue}{\texttt{\# Idea Generation}}}
\STATE \texttt{idea = scientist.\textcolor{Green}{think}(intent)}

\STATE \texttt{\textcolor{CadetBlue}{\texttt{\# Code Experiment}}}
\STATE \texttt{status, exp\_dir = scientist.\textcolor{Green}{code}(idea)}

\STATE \texttt{if status:}
\STATE \hspace{1em}\texttt{\textcolor{CadetBlue}{\texttt{\# Paper Writing}}}
\STATE \hspace{1em}\texttt{pdf\_path = scientist.\textcolor{Green}{write}(idea, exp\_dir)}
\STATE \hspace{1em}\texttt{\textcolor{CadetBlue}{\texttt{\# Paper Review}}}
\STATE \hspace{1em}\texttt{review = scientist.\textcolor{Green}{review}(pdf\_path)}
\end{algorithmic}
\end{algorithm}

\begin{figure}[!t]
    \centering
    \includegraphics[width=\linewidth]{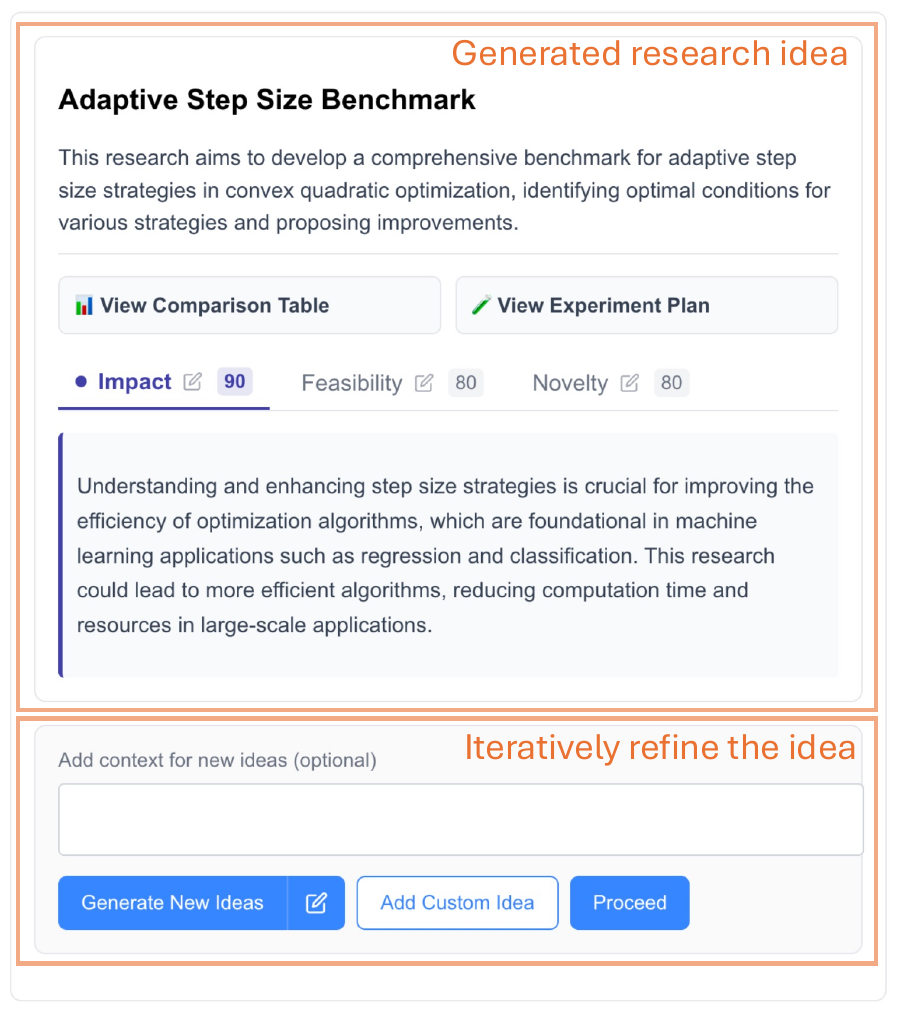}
    \caption{\textbf{Example of iterative interaction within the thinking stage.} The upper box shows a research idea (including contents, scores, tables, and experimental plans) generated by the Thinker. The lower box allows users to provide custom instructions to refine.}
    \label{fig:ui-feature1}
    \vspace{-2mm}
\end{figure}

\section{\modelname User Interface}

To further enhance the usability of \modelname, we develop a user interface that leverages the \modelname Python package to build its backend. The interactive and modular nature of the \modelname design lends itself to an intuitive UI design. In the UI, each workflow stage is presented on a dedicated page, arranged sequentially. Details about the UI design are available at Appendix~\S\ref{appendix:ui-details}.

\vspace{1mm}
\xhdr{Iterative interaction within one stage}  
This feature arises from the iterative agent design introduced in Section~\S\ref{subsection:workflow-components}. Since each core workflow component supports iterative refinement, \modelname allows fine-grained user inputs to be injected during the agent’s reflection process. As shown in Figure~\ref{fig:ui-feature1}, human researchers can continuously add or adjust intents, prompting the system to refine and regenerate ideas accordingly.

\vspace{1mm}
\xhdr{Tabular-based interaction across stages}  
This functionality is built upon the tabular-based communication mechanism discussed in Section~\S\ref{subsection:feature-components}. In \modelname, generated ideas and experimental plans are organized into structured tables, making them easy to interpret and edit. Figure~\ref{fig:ui-feature2} presents an example of a table generated by \modelname for novelty comparison. The tabular format highlights key differences between generated ideas and prior work, allowing human researchers to clearly understand, compare, and modify content as needed.

\begin{table}[t]
\centering
\small
\caption{\textbf{Ablation study on tool using.} We compare the paper generation quality of \modelname with and without tool usage. LLM evaluation uses GPT-4o as the evaluator. ``Writing'' refers to writing quality and ``idea'' refers to idea quality.}
\label{tab:ablation_notbio_tool}
\begin{tabular}{lcccc}
\toprule
\multirow{2}{*}{\textbf{Method}} 
& \multicolumn{2}{c}{\textbf{LLM Evaluation}} 
& \multicolumn{2}{c}{\textbf{Human Evaluation}} \\
\cmidrule(lr){2-3} \cmidrule(lr){4-5}
& \multicolumn{1}{c}{Writing} & \multicolumn{1}{c}{Idea} 
& \multicolumn{1}{c}{Writing} & \multicolumn{1}{c}{Idea} \\
\midrule
w/ tool  & \textbf{4.06} & \textbf{3.82} & \textbf{3.93} & \textbf{3.83} \\
w/o tool & 3.93 & 3.78 & 3.80 & 3.68 \\
\bottomrule
\end{tabular}
\vspace{-5mm}
\end{table}

\vspace{-1mm}
\section{Evaluation Results}
\vspace{-1mm}

In addition to qualitatively analyzing the Python package and the user interface of \modelname, we conduct quantitative evaluations to verify that our agentic framework—designed for enhanced interactivity, extensibility, and controllability—maintains the quality of generated papers.

\vspace{-1mm}
\subsection{Evaluation Settings}

\xhdr{Model settings} We use \texttt{gpt-4o-mini} as the backbone model for both \modelname and the Agent Laboratory to ensure a fair comparison between the two agent frameworks.

\vspace{1mm}
\xhdr{Data settings} We evaluate two categories of tasks.  
(1) \textit{In-distribution tasks:} We randomly sample 20 machine learning-related ideas from \citet{si2024can}, using their idea titles as user intent inputs.  
(2) \textit{Out-of-distribution tasks:} To test the safety and robustness of \modelname, we randomly sample 20 biology-related potentially unsafe tasks (\textit{e.g.}, DNA synthesis for synthetic genomes) from SciSafetyBench~\citep{zhu2025safescientistriskawarescientificdiscoveries} and use them as input intents for both frameworks.

\vspace{1mm}
\xhdr{Evaluation settings} We conduct both automated and human evaluations, focusing on two key dimensions: (1) \textbf{Writing quality} assesses the clarity of writing, appropriate use of citations, and richness of content. (2) \textbf{Idea quality} measures the research value and usefulness behind the generated papers. To ensure consistency, we use the same evaluation rubric for both LLM-based and human assessments: the rubric is provided as a prompt to the LLM and as written guidelines to human annotators. Human evaluations are performed by annotators with relevant research backgrounds, who assess the quality of the generated papers from both frameworks. Rubrics are in Appendix~\S\ref{human-evaluation-details}.

\vspace{-1mm}
\subsection{Evaluation Results}
\vspace{-1mm}

\begin{figure}[t]
    \centering
    \includegraphics[width=\linewidth]{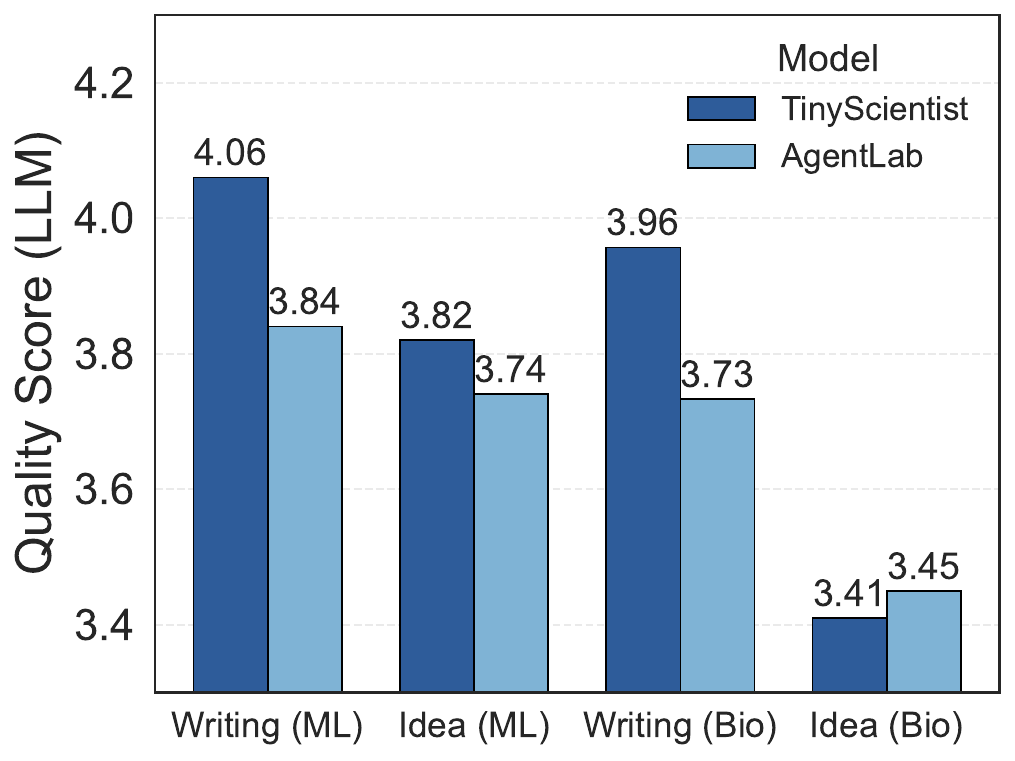}
    
    \caption{\textbf{LLM-based evaluation results}. We report 5-scale quality scores assigned by LLM judges (with GPT-4o) for the generated paper outputs. The evaluation covers both the biological and ML domains, and each includes (1) writing quality and (2) idea quality.}
    \label{fig:llm-quality-eval}
    \vspace{-3mm}
\end{figure}

\begin{figure}[t]
    \centering
    \includegraphics[width=\linewidth]{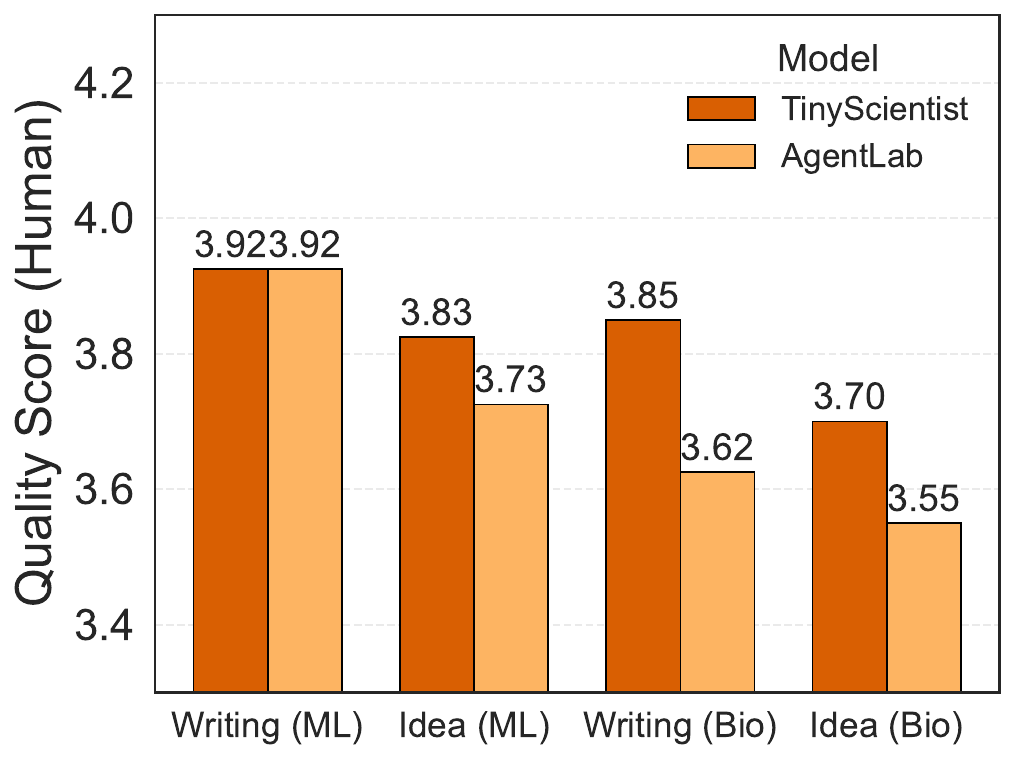}
    
    \caption{\textbf{Human evaluation results}. We report 5-scale quality scores assigned by human annotators for the generated paper outputs. The evaluations are conducted under the same settings and rubrics as those used for the LLM-as-the-judge evaluation.}
    \label{fig:human-quality-eval}
    \vspace{-3mm}
\end{figure}

\vspace{1mm}
\xhdr{\modelname achieves better generation quality to Agent Laboratory}
As shown in Figure~\ref{fig:llm-quality-eval} and Figure~\ref{fig:human-quality-eval}, the writing quality and idea quality of papers generated by \modelname are better—or at least comparable under all cases—than those produced by the Agent Laboratory. These results are consistently supported by both LLM-based and human evaluations. Notably, the largest improvement appears in the biological writing domain, with a 0.23-point increase under both evaluation settings. We attribute this improvement to richer tool using and more structured prompting, which leads to more informative generations. However, for out-of-distribution biological cases, both \modelname and the Agent Laboratory exhibit lower performance, likely because the prompt design is primarily tailored to the ML domain, making the generated paper less valuable.

\vspace{1mm}
\xhdr{Tool usage in \modelname improves paper generation quality}
We conduct a focused evaluation to examine the effect of tool usage within \modelname on ML-related tasks. As shown in Table~\ref{tab:ablation_notbio_tool}, both human and LLM-based evaluations show clear improvements when paper generation is augmented with tools such as the paper searcher, with more than a 0.1-point gain in writing quality and around a 0.05-point gain in idea quality. These results demonstrate the effectiveness of the tool-augmented workflow in \modelname.

\vspace{1mm}
\xhdr{Checker in \modelname blocks unsafe user intents effectively}
To evaluate the safety controllability of \modelname, we conduct evaluations under out-of-distribution tasks in biological domains with potential ethical risk. Among the total 20 tasks for generation, 18 tasks were blocked at the thinker stage, while the remaining 2 were flagged with warnings at the further workflow stages. These results demonstrate that the checker enables \modelname to effectively prevent the development of potentially harmful research.

\vspace{1mm}
\xhdr{Checker in \modelname adaptively adjusts the workflow based on budget}
For the budget controllability of \modelname, we test under different budgets for different backbone models. The checker estimates the cost and dynamically adjusts the number of \textit{reflection} steps within each stage of the \modelname (thinker, writer, coder, reviewer, checker). Each stage has its own estimated cost ratio under the overall budget. Furthermore, if the actual cost exceeds the budget due to estimation errors, the checker terminates the process early to prevent overspending. This mechanism ensures strict budget control throughout the paper generation process.

\vspace{-2mm}
\section{Conclusion}
\vspace{-2mm}
In this work, we present \modelname, a lightweight framework that prioritizes simplicity and usability to democratize the development of research agents. To enhance accessibility, we developed an easy-to-use Python package and a highly interactive web demonstration for it. We believe that \modelname would help lower the barrier for both researchers and developers to enter the field of automatic research, encouraging more people to contribute to its development.

\vspace{-1mm}
\section*{Ethical Statements and Broader Impacts}
\vspace{-1mm}
The development of \modelname targets for extensive and interactive Automated research workflows carries inherent ethical risks, including the potential for misuse, unintended harm, or malicious exploitation. To mitigate these concerns, \modelname incorporates multiple safeguards: (1) configurable budget and safety controls, (2) automated watermarking on generated papers to clearly indicate AI involvement, and (3) an interactive user interface that supports real-time human oversight and intervention. We emphasize that \modelname is designed to augment—not replace—human researchers. Its goal is to accelerate scientific discovery through transparent, collaborative human-AI workflows, not to enable fully autonomous research without accountability. To further uphold ethical standards, we openly release \modelname as a Python library with user-friendly interfaces, making advanced research capabilities accessible to a broader community while maintaining transparency and control.

Regarding generated data, we acknowledge that some AI-generated artifacts (\textit{e.g.}, papers or figures) may closely resemble human-written content. To prevent misuse, all generated outputs are clearly watermarked and not intended for direct use in academic publishing without human verification. This aligns with best practices for responsible research and ensures that the system and its outputs are used in ethically appropriate ways.

\vspace{-1mm}
\section*{Limitations}
\vspace{-1mm}
There are two main limitations for our work:

\vspace{1mm}
\xhdr{Compilation stability} While our system performs robustly in most scenarios, compilation failures occasionally arise due to inconsistencies in \textsc{LaTeX} formatting, which causes issues such as references missing or line misalignment. Improved context sanitizer and format-control generation are needed to ensure stability across all outputs. More robust methods for \textsc{LaTeX} error/warning correction should be included as part of the output formatter in the future version of \modelname.

\vspace{1mm}
\xhdr{Diagram quality and informativeness} Although our system can generate diagrams for key sections such as the Introduction and Method, the visual quality and informativeness of these figures remain limited. Improving visual consistency and content-grounding in diagram generation would significantly enhance the entire paper's clarity.

\section*{Acknowledgments}
We sincerely appreciate the support from the Amazon grant
funding project \#120359, ``GRAG: Enhance RAG Applications with Graph-structured Knowledge'', and Meta gift
funding project ``PERM: Toward Parameter Efficient Foundation Models for Recommenders''.

\bibliography{custom}

\newpage

\appendix
\onecolumn
\section{User Interface Details}
\label{appendix:ui-details}
\begin{figure}[H]
    \centering
    \includegraphics[width=0.8\textwidth]{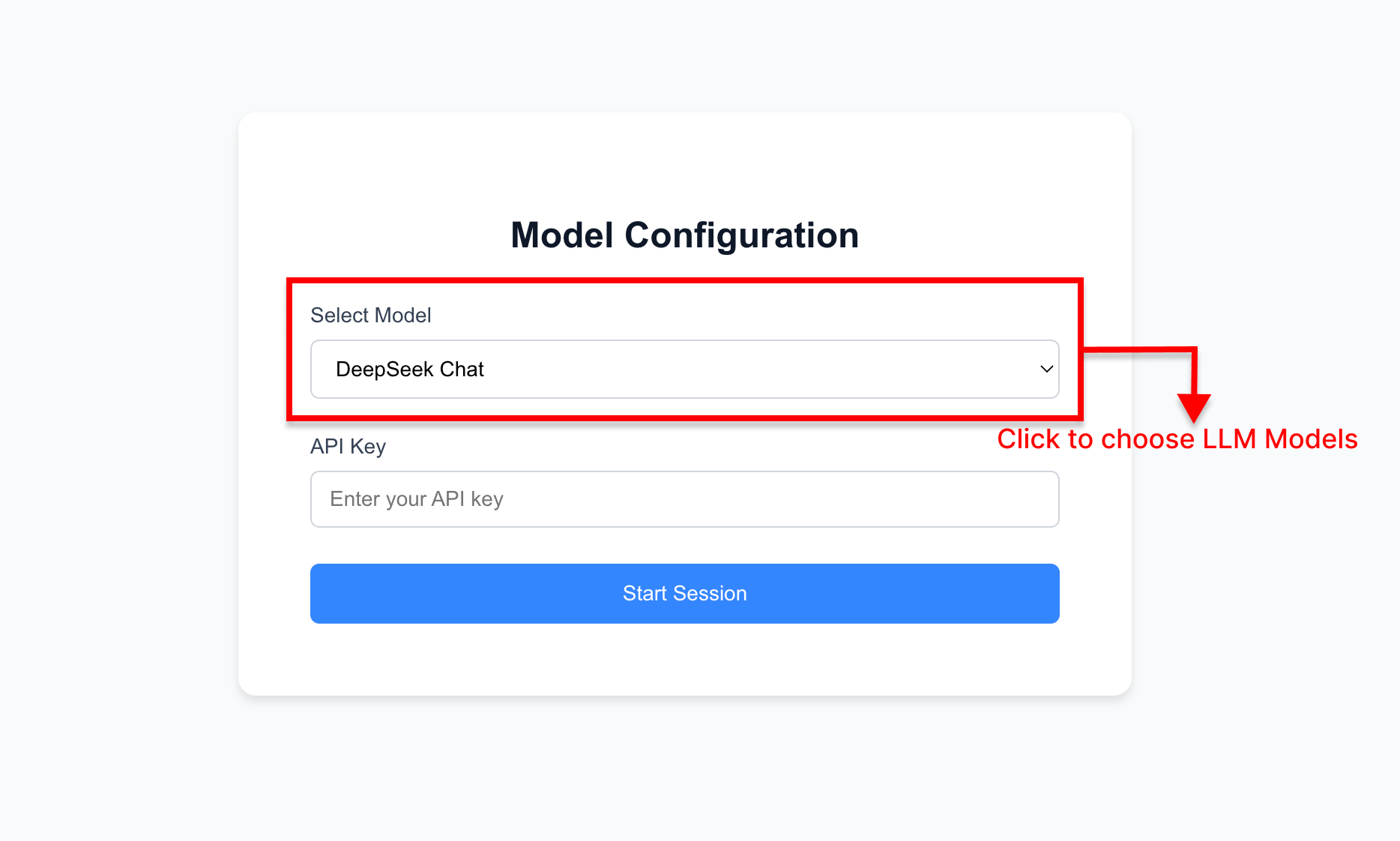} 
    \caption{\textbf{Screenshot for configuration input page}. Users are guided to select the LLM model, provide their API key, and click \emph{Start Session} to proceed to the next stage of idea generation. We would not save the user's API key to our server.}
    \label{fig:page-1}
\end{figure}

\begin{figure}[H]
    \centering
    \includegraphics[width=0.8\textwidth]{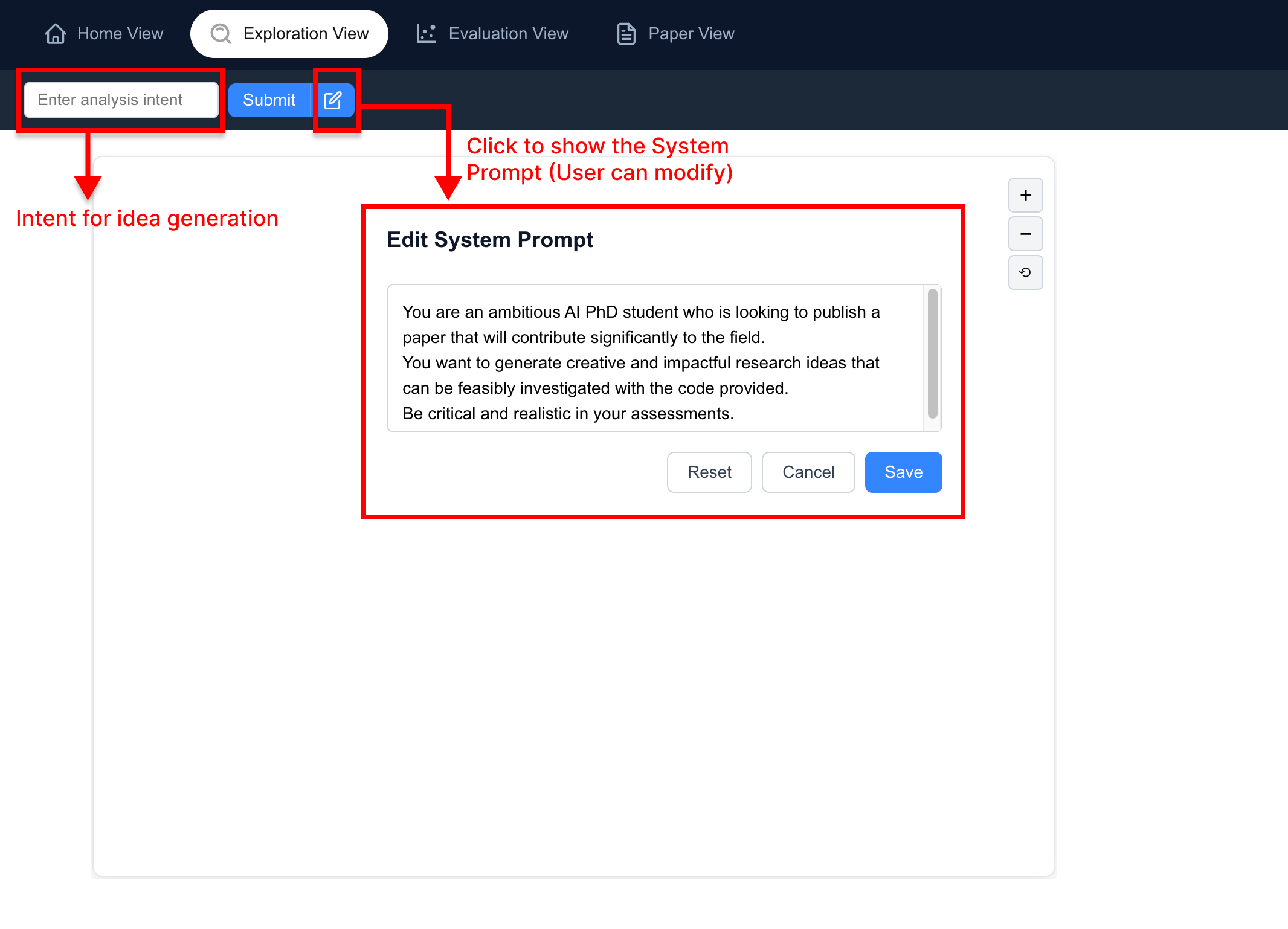} 
    \caption{\textbf{Screenshot for intent input page}. Users enter their research intent and click \emph{Submit} to generate three candidate ideas. The icon next to \emph{Submit} reveals the current system prompt, which can be modified to better align with the research intent.}
    \label{fig:page-2}
\end{figure}

\begin{figure}[H]
    \centering
    \begin{subfigure}[b]{0.85\textwidth}
        \centering
        \includegraphics[width=\textwidth]{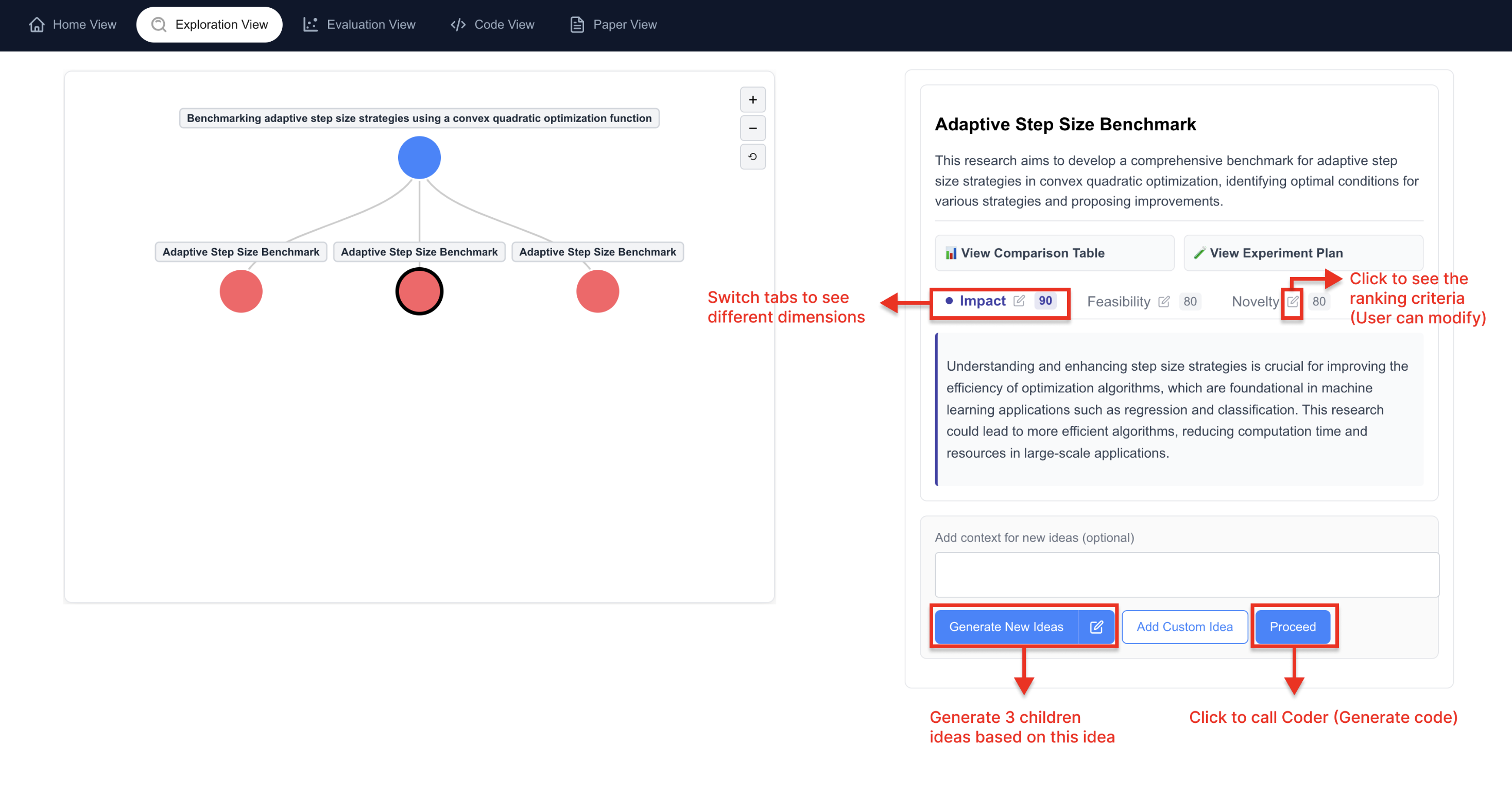}
        \vspace{-2mm}
        \caption{Main idea view.}
        \label{fig:view_main}
    \end{subfigure}
    \vfill
    \begin{subfigure}[b]{0.85\textwidth}
        \centering
        \includegraphics[width=\textwidth]{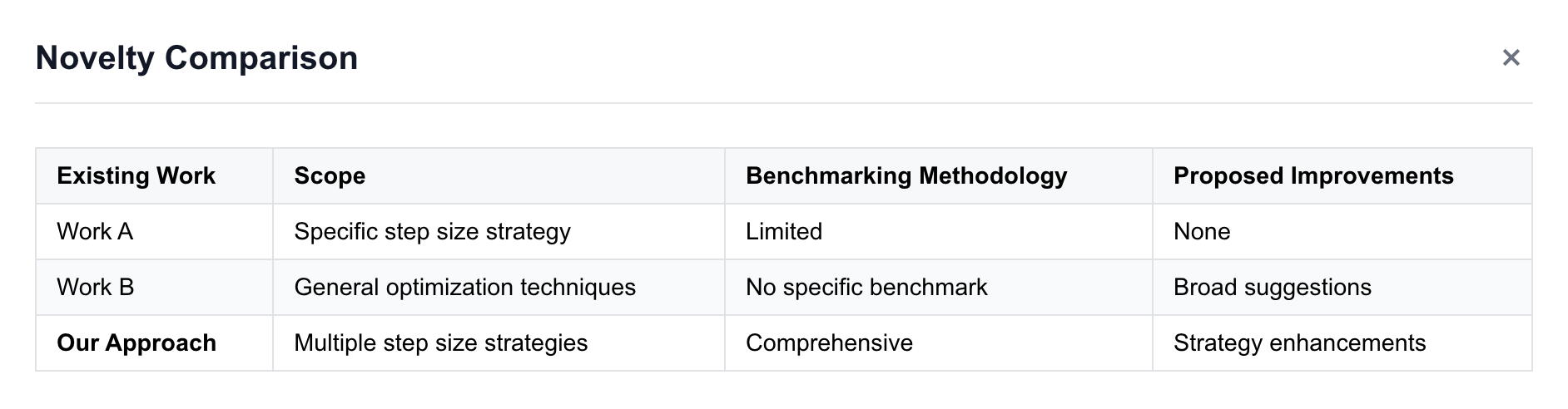}
        \caption{Comparison table.}
        \label{fig:idea_table}
    \end{subfigure}
    \hfill
    \begin{subfigure}[b]{0.85\textwidth}
        \centering
        \includegraphics[width=\textwidth]{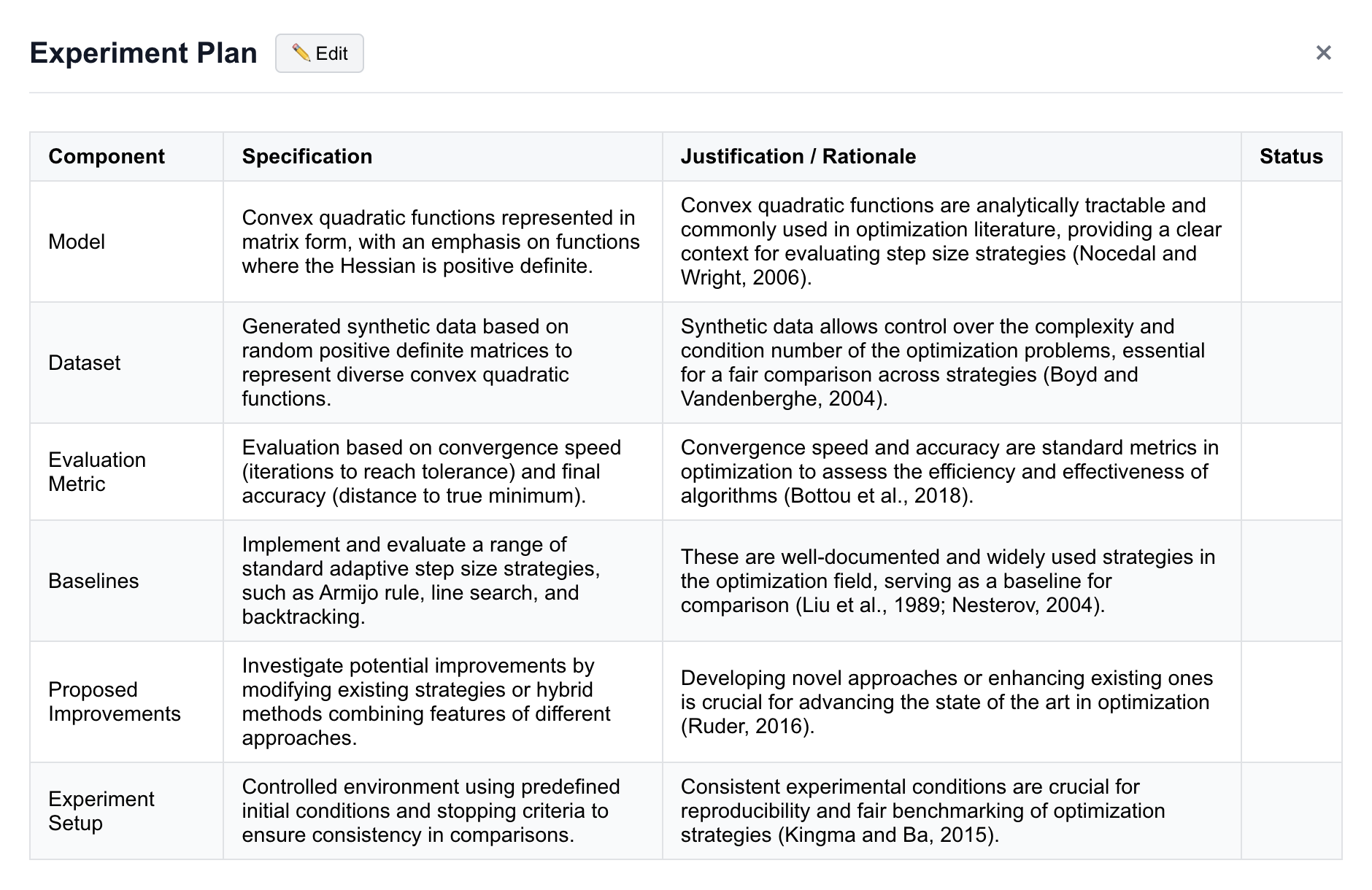}
        \caption{Experiment plan table.}
        \label{fig:exp_table}
    \end{subfigure}

    \caption{\textbf{Screenshot for idea viewing page}. (a) The main idea view, where users can explore ideas by clicking on nodes. (b) Comparison table, accessed by clicking \emph{View Comparison Table}. (c) The experiment table, accessed by clicking \emph{View Experiment Table}.}
    \label{fig:page-3}
\end{figure}

\begin{figure}
    \centering
    \includegraphics[width=0.8\textwidth]{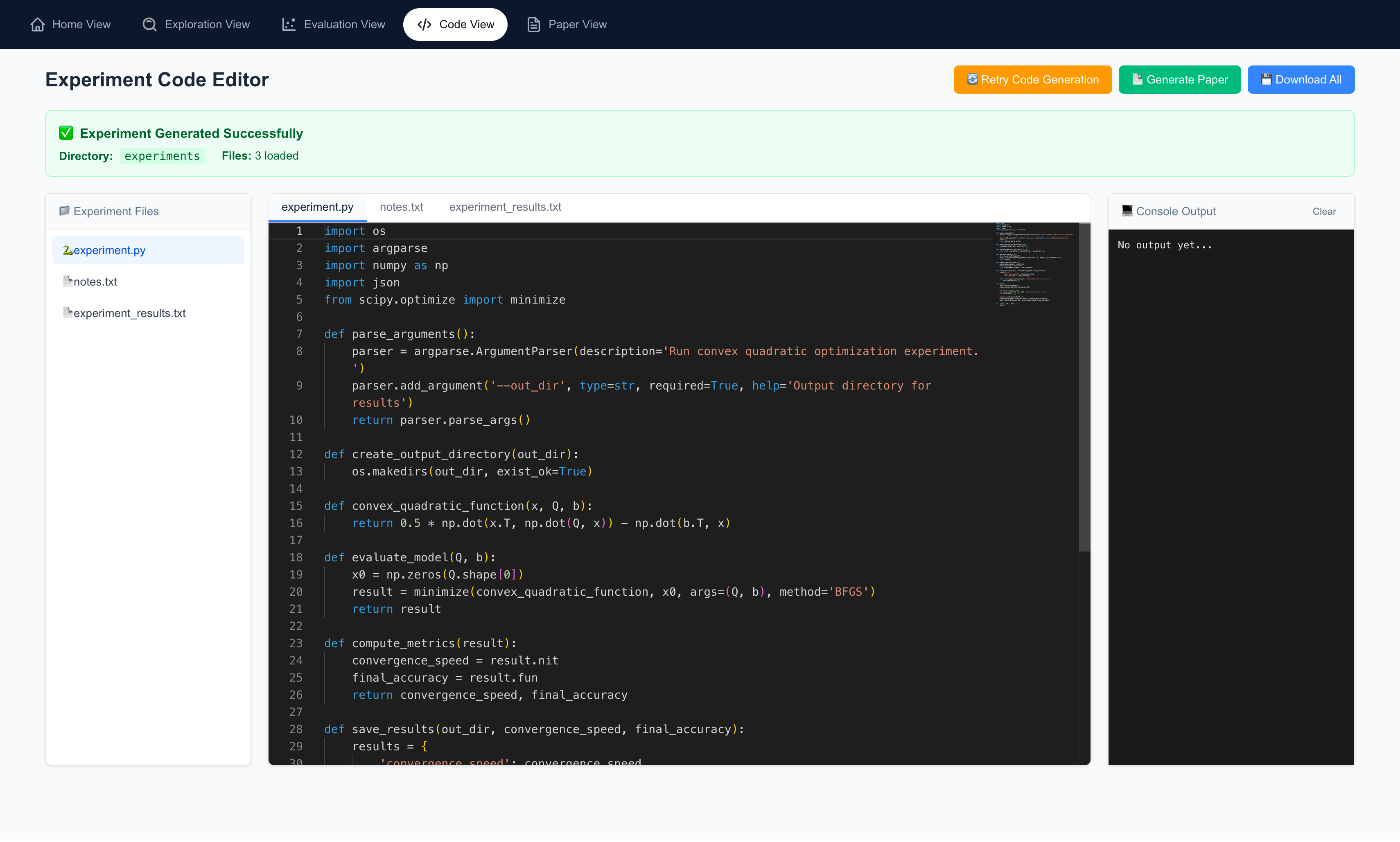}
    \caption{\textbf{Screenshot for code viewing page}. The code view is displayed when the Coder module is invoked. Users may click \emph{Download All} to export the generated experiment files as a zip archive, or \emph{Generate Paper} to call the Writer module to draft a research paper.}
    \label{fig:page-4}
\end{figure}

\begin{figure}[H]
    \centering
    \begin{subfigure}[b]{0.85\textwidth}
        \centering
        \includegraphics[width=\textwidth]{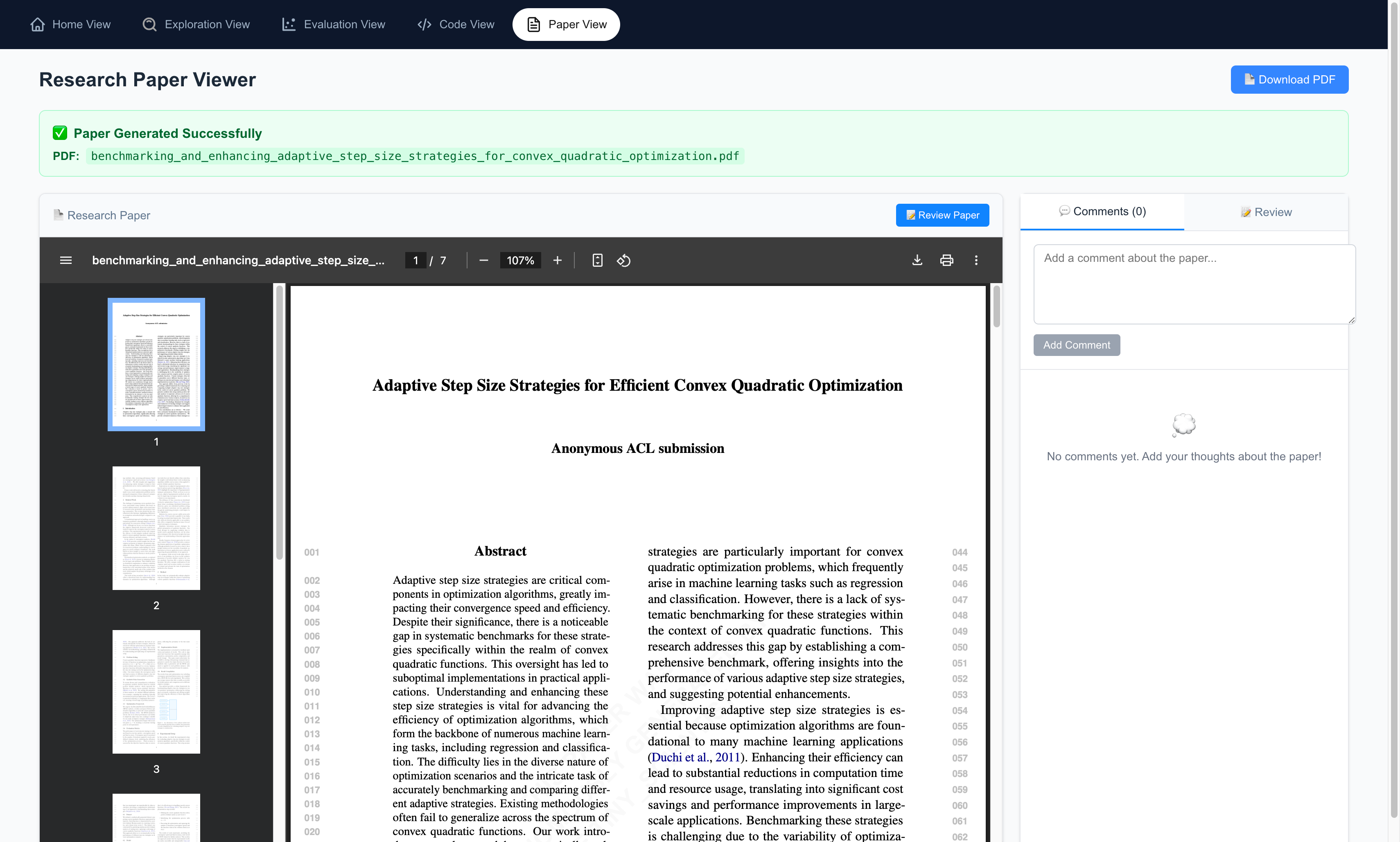}
        \caption{Paper PDF preview}
        \label{fig:evalView}
    \end{subfigure}
    \vfill
    \begin{subfigure}[b]{0.85\textwidth}
        \centering
        \includegraphics[width=\textwidth]{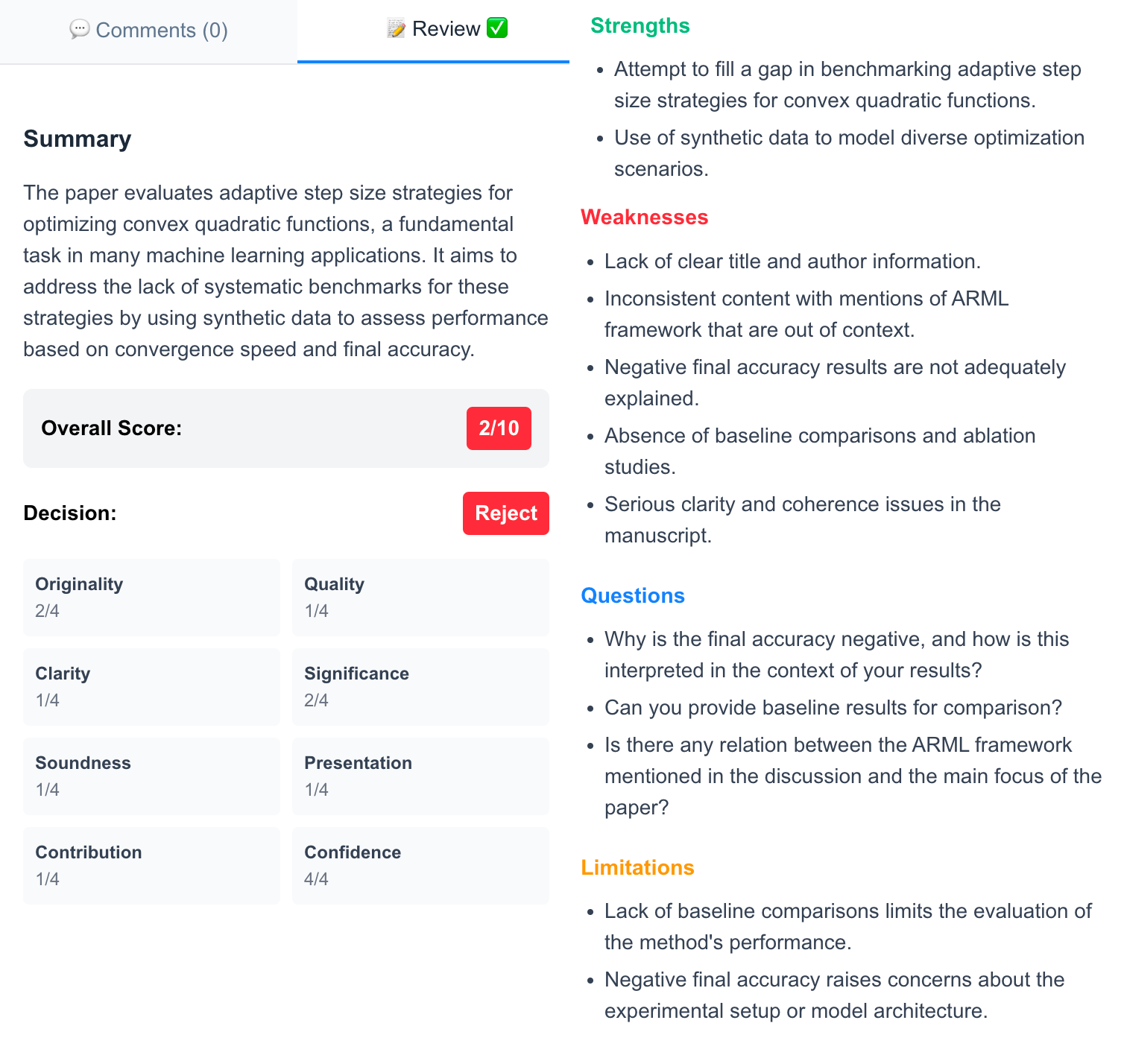}
        \caption{Paper review}
        \label{fig:evalDrag}
    \end{subfigure}
    \caption{\textbf{Screenshot for the paper viewing page}. (a) The paper PDF preview, where users can preview the generated paper PDF and click \emph{Download PDF} to save it, or the user can also click \emph{Review Paper} to invoke the paper reviewing stage. (b) The paper review, where LLM-based reviewers are utilized to generate a paper review for the generated paper PDF.}
    \label{fig:page-5}
\end{figure}

In this section, we provide a step-by-step guide to the \modelname user interface, which consists of five main pages. The first is the \emph{configuration input page} (Figure~\ref{fig:page-1}), where users provide API keys and select backbone models. After configuration, users proceed to the \emph{intent input page} (Figure~\ref{fig:page-2}) to describe their research intent. Next, the \emph{idea viewing page} (Figure~\ref{fig:page-3}) presents tree-structured idea generation results, together with tabular descriptions for novelty comparison and experiment plans. On this page, it also allows users to iteratively refine ideas by giving text-based feedback. Once an idea is confirmed, the interface moves to the \emph{code viewing page} (Figure~\ref{fig:page-4}), where users can download the generated file structure. Finally, the \emph{paper viewing page} (Figure~\ref{fig:page-5}) provides a PDF preview of the generated paper along with reviews generated from the review component of \modelname.

\section{Prompting Details}
In this section, we provide the details about the prompt used for each workflow stage in \modelname.
\label{sec:appendix: prompt technical details}
\subsection{Thinker prompt}
We present the complete set of prompts used in the \modelname thinker module. The system prompt establishes the agent role as a research scientist and defines core guidelines for idea generation (Table~\ref{tab:thinker_system_prompt}). The idea generation and refinement prompts handle the creation and modification of research ideas through structured approaches to problem identification and solution development (Tables~\ref{tab:idea_generation_prompt} and \ref{tab:idea_modification_prompt}). The evaluation prompts provide a comprehensive assessment mechanism for evaluating research ideas (Table~\ref{tab:idea_evaluation_prompt} and ~\ref{tab:novelty_evaluation_prompt}).

\subsection{Coder prompt}
We present the complete set of prompts used in the \modelname coder module. The system prompt establishes the agent role as a research scientist and defines core guidelines for experiment implementation (Table~\ref{tab:coder_system_prompt}). The code execution and error handling prompts handle the refinement and re-plan for the experiment in different scenarios (Tables~\ref{tab:error_handling_prompt}), and the experiment format prompt controls the output format of essential experiment details (Tables~\ref{tab:experiment_format_prompt}).

\subsection{Writer prompt}
We present the complete set of prompts 
used in the \modelname writer module. The system prompt establishes the agent role as a research scientist and defines core guidelines for each step of paper writing (Table~\ref{tab:writer_system_prompt}). Section instructions prompts provide details, tips, and instructions for section writing (Table~\ref{tab:section_tips_prompt}, Table~\ref{tab:abstract_instruction_prompt}). The citation management prompts are responsible for citation fetching and embedding (Table~\ref{tab:citation_system_prompt}, Table~\ref{tab:citation_collection_prompt}, Table~\ref{tab:citation_embedding_prompt}). If there is error in rendering PDF, we utilize the refinement prompt to solve (Table~\ref{write-refinement}).

\subsection{Reviewer prompt}
We present the complete set of prompts used in the \modelname reviewer module. The system prompt establishes the agent role as a research scientist and defines core guidelines for paper review (Table~\ref{tab:reviewer_system_prompt}). Prompts Review Format \& Refinement are responsible for providing detail review guidelines and format control (Table~\ref{tab:reviewer_template_prompt}, Table~\ref{tab:reviewer_reflection_prompt}).

\section{Human Evaluation Details}
\label{human-evaluation-details}
We provide the detailed quality evaluation rubrics for both LLM-based evaluation and human evaluation in Table~\ref{tab:writing_quality_evaluation_rubrics} (writing quality) and Table~\ref{tab:idea_quality_evaluation_rubrics} (idea quality).

\section{Case Study}
\label{case-study}
\begin{figure}[H]
    \centering
    \includegraphics[width=0.5\textwidth]{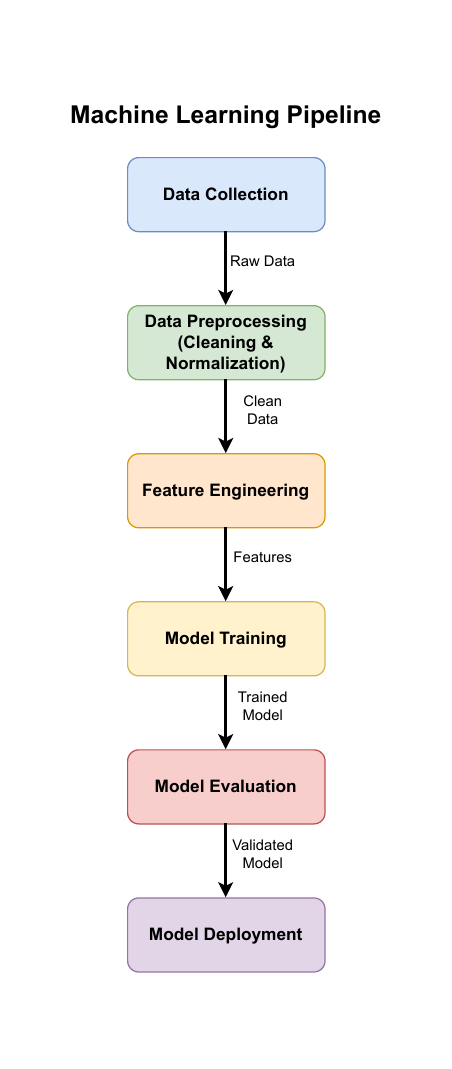}
    \caption{\textbf{Case study on drawer}. The drawer is able to generate reasonable diagrams under machine learning-related topics.}
    \label{fig:diagram}
\end{figure}

We first present a case study of the checker as part of the feature component, with its output shown in Table~\ref{tab:safechecker_result}.
For MCPClient, we provide a case study of the drawer, a commonly useful tool. The corresponding input is shown in Table~\ref{tab:drawio_code_generator_prompt}, and the generated output is illustrated in Figure~\ref{fig:diagram}.

\clearpage

\clearpage

\clearpage

\clearpage
\begin{table}
\centering
\vspace*{\fill}
\caption{\textbf{Thinker system prompt}.}
\label{tab:thinker_system_prompt}
\small
\begin{tabular}{|p{0.95\textwidth}|}
\hline
\textbf{Thinker System Prompt} \\
\hline
\begin{minipage}[t]{0.92\textwidth}
\vspace{0.1em}
\begin{verbatim}
IDEA_SYSTEM_PROMPT: 
  You are an ambitious AI PhD student who is looking 
  to publish a paper that will contribute significantly to the field.
  You want to generate creative and impactful 
  research ideas that can be feasibly investigated with the code provided.
  Be critical and realistic in your assessments.
EVALUATION_SYSTEM_PROMPT: 
  You are an expert research reviewer who evaluates scientific ideas with rigor and fairness.
  Your role is to comparatively evaluate multiple 
  research ideas and rank them based on their 
  feasibility, novelty, impact, and alignment with the original research intent.
  Be thoughtful, objective, and provide clear justifications for your rankings.
NOVELTY_SYSTEM_PROMPT: |
  You are an ambitious AI PhD student who is 
  looking to publish a paper that will contribute significantly to the field.
  You have an idea and you want to check if it is novel or not. 
  I.e., not overlapping significantly with existing literature or already well explored.
  Be a harsh critic for novelty, ensure there is a 
  sufficient contribution in the idea for a new conference or workshop paper.
  You are analyzing search results to determine if 
  your idea has already been explored in existing literature.
  Decide a paper idea is novel if after sufficient 
  searching, you have not found a paper that significantly overlaps with your idea.
  Decide a paper idea is not novel if you have 
  found a paper that significantly overlaps with your idea.
ETHICAL_SYSTEM_PROMPT: >
  You are an expert AI research ethics advisor. 
  Your role is to review research ideas and ensure they align with scientific ethical standards. 
  You help researchers enhance their ideas to be 
  more ethical, beneficial, and responsible while maintaining their scientific value.
  Focus on identifying potential risks and 
  suggesting constructive improvements that make research more ethically sound.

\end{verbatim}
\vspace{0.1em}
\end{minipage} \\
\hline
\end{tabular}
\end{table}

\clearpage
\begin{table}
\centering
\caption{\textbf{Thinker idea generation prompt}.}
\label{tab:idea_generation_prompt}
\footnotesize
\begin{tabular}{|p{0.95\textwidth}|}
\hline
\textbf{Thinker Idea Generation Prompt} \\
\hline
\begin{minipage}[t]{0.92\textwidth}
\vspace{0.1em}
\begin{verbatim}
IDEA_GENERATION_PROMPT: 
  Generate a creative and impactful research idea based on the following intent:
  ```
  {intent}
  ```
  ```
  {pdf_section}
  ```
  Additionally, based on recent literature, here 
  are some related works that might inform your next idea:
  ```
  {related_works_string}
  ```
  Based on the above, come up with the next 
  impactful and creative research idea that addresses the following questions:
  1. What is the problem?
    - Provide a comprehensive description of the 
    research problem, including background, current challenges, and why the issue persists.
    - Include citations where relevant. All 
    citations should be in parentheses (e.g., (Workowski & Bolan, 2015)).
    - Make sure this problem statement directly addresses the original intent.
  2. Why is it interesting and important?
    - Explain in detail why the problem is 
    interesting and important. Support your claims with references from recent literature.
    - Connect the importance back to the original intent.
  3. Why is it hard?
    - Analyze the inherent challenges of the 
    problem and explain why naive approaches have failed, citing previous studies.
    - Discuss why this problem remains difficult in the context of the original intent.
  4. Why hasn't it been solved before?
    - Clearly describe how your idea differs from 
    existing solutions. Highlight innovative aspects and include comparative citations.
    - Explain why existing approaches from the related works don't fully address the intent.
  5. What are the key components of my approach and results?
    - Outline your proposed methodology.
    - Explain how your approach specifically addresses the original intent.
  Respond in the following format:

  THOUGHT:
  <THOUGHT>

  NEW IDEA JSON:
  ```json
  <JSON>
  ```

  Be cautious and realistic on your ratings.
  This JSON will be automatically parsed, so ensure the format is precise.
  You will have {num_reflections} rounds to iterate on the idea, but do not need to use them all.

  Completed ideas have an additional "Score" field 
  which indicates the assessment by an expert ML reviewer.
  This is on a standard 1-10 ML conference scale.
  Scores of 0 indicate the idea failed either during experimentation, writeup or reviewing.

\end{verbatim}
\vspace{0.1em}
\end{minipage} \\
\hline
\end{tabular}
\end{table}

\clearpage
\begin{table}
\centering
\caption{\textbf{Thinker idea modification prompt}.}
\label{tab:idea_modification_prompt}
\footnotesize
\begin{tabular}{|p{0.95\textwidth}|}
\hline
\textbf{Thinker Idea Modification Prompt} \\
\hline
\begin{minipage}[t]{0.92\textwidth}
\vspace{0.1em}
\begin{verbatim}
IDEA_MODIFICATION_PROMPT: 
  Given a research idea and a set of requested modifications, generate a modified version of the idea.

  ORIGINAL RESEARCH IDEA:
  ```
  {idea}
  ```

  REQUESTED MODIFICATIONS:
  ```
  {modifications}
  ```

  RESEARCH INTENT:
  ```
  {intent}
  ```
  Carefully consider how to preserve the core 
  strengths of the original idea while enhancing it 
  according to the requested modifications. Ensure 
  the modified idea maintains strong alignment with the original research intent.

  For each modification request, adjust the 
  corresponding aspect (Novelty, Feasibility, or 
  Impact) by emphasizing or de-emphasizing relevant characteristics.

  Respond in the following format:

  THOUGHT:
  <THOUGHT>

  MODIFIED IDEA JSON:
  ```json
  <JSON>
  ```

  In <THOUGHT>, explain your reasoning for the modifications and how they enhance the idea.
  In <JSON>, provide the modified idea with the 
  same structure as the original, including all original fields.
\end{verbatim}
\vspace{0.1em}
\end{minipage} \\
\hline
\end{tabular}
\end{table}

\clearpage
\begin{table}
\centering
\caption{\textbf{Thinker idea evaluation prompt}.}
\label{tab:idea_evaluation_prompt}
\footnotesize
\begin{tabular}{|p{0.95\textwidth}|}
\hline
\textbf{Thinker Idea evaluation prompt} \\
\hline
\begin{minipage}[t]{0.92\textwidth}
\vspace{0.1em}
\begin{verbatim}
IDEA_EVALUATION_PROMPT: 
  You are tasked with evaluating and scoring 
  multiple research ideas generated for the following research intent:

  RESEARCH INTENT:
  ```
  {intent}
  ```
  RESEARCH IDEAS TO EVALUATE:
  ```
  {ideas}
  ```
  Please evaluate these ideas comparatively across three key dimensions:

  **NOVELTY DIMENSION**
  {novelty_criteria}
  **FEASIBILITY DIMENSION**
  {feasibility_criteria}
  **IMPACT DIMENSION**
  {impact_criteria}

  CRITICAL REQUIREMENTS:
  1. For EACH idea, you MUST provide three separate rating fields that MUST follow this format:
    - "FeasibilityScore": A number from 0 to 100, where 100 is most feasible
    - "NoveltyScore": A number from 0 to 100, where 100 is most novel
    - "ImpactScore": A number from 0 to 100, where 100 is highest impact

  2. For EACH idea, also provide a brief reasoning for each score:
    - "NoveltyReason": Brief explanation (1-2 
    sentences) of why this idea received its novelty score
    - "FeasibilityReason": Brief explanation (1-2 
    sentences) of why this idea received its feasibility score
    - "ImpactReason": Brief explanation (1-2 
    sentences) of why this idea received its impact score
  These three scores must be completely separate 
  and independent from each other. For example, the 
  idea with the highest impact score might have a low feasibility score.

  Respond in the following format:

  COMPARATIVE ANALYSIS:
  <ANALYSIS>
  EVALUATION JSON:
  ```json
  <JSON>
  ```

  In <ANALYSIS>, provide a thoughtful comparative analysis discussing the trade-offs between ideas.
  In <JSON>, provide the evaluation results in JSON format with the following structure:
  "scored_ideas": A list of scored idea objects, each containing:
  - "Title": The EXACT original title of the idea 
  as provided in the input JSON - DO NOT MODIFY OR CHANGE THE TITLE IN ANY WAY
  - "FeasibilityScore": A number from 0 to 100, scoring feasibility
  - "NoveltyScore": A number from 0 to 100, scoring novelty
  - "ImpactScore": A number from 0 to 100, scoring impact
  - "NoveltyReason": Explanation of the novelty score
  - "FeasibilityReason": Explanation of the feasibility score
  - "ImpactReason": Explanation of the impact score

  CRITICAL: You MUST preserve the exact original 
  titles from the input. Do not change, modify, or improve the titles in any way.
  Ensure your evaluation is fair, comprehensive, 
  and based solely on the scientific and practical merits of each idea.
\end{verbatim}
\vspace{0.1em}
\end{minipage} \\
\hline
\end{tabular}
\end{table}

\clearpage
\begin{table}
\centering
\caption{\textbf{Thinker idea novelty evaluation prompt}.}
\label{tab:novelty_evaluation_prompt}
\footnotesize
\begin{tabular}{|p{0.95\textwidth}|}
\hline
\textbf{Thinker Idea Novelty evaluation prompt} \\
\hline
\begin{minipage}[t]{0.92\textwidth}
\vspace{0.1em}
\begin{verbatim}
NOVELTY_PROMPT: 
  Round {current_round}/{num_rounds}.
  You are assessing the novelty of the following 
  research idea in the context of the original intent:

  ORIGINAL INTENT:
  ```
  {intent}
  ```

  CURRENT IDEA:
  ```
  {idea}
  ```

  SEARCH RESULTS FROM PREVIOUS QUERY:
  ```
  {last_query_results}
  ```

  Respond in the following format:

  THOUGHT:
  <THOUGHT>

  DECISION:
  <DECISION>

  In <THOUGHT>, carefully analyze the idea's novelty by:
  1. First explicitly assess how well the idea aligns with the original intent
  2. Compare the idea against the search results to identify similarities and differences
  3. Determine if any existing work already implements the core approach for the same intent
  4. Consider if the idea offers meaningful innovation beyond existing approaches
  5. Assess whether minor variations from existing work constitute sufficient novelty

  In <DECISION>, write either:
  - "NOVELTY CHECK: CONTINUE" if you need more 
  information to make a decision. In this case, explain what specific information you need.
  - "NOVELTY CHECK: NOVEL" if you've determined the idea is novel. Briefly explain why.
  - "NOVELTY CHECK: NOT NOVEL" if you've determined 
  the idea is not novel. Briefly explain why and 
  cite the specific paper(s) that demonstrate lack of novelty.

\end{verbatim}
\vspace{0.1em}
\end{minipage} \\
\hline
\end{tabular}
\end{table}

\clearpage
\begin{table}
\centering
\caption{\textbf{Coder system prompt}.}
\label{tab:coder_system_prompt}
\footnotesize
\begin{tabular}{|p{0.95\textwidth}|}
\hline
\textbf{Coder System Prompt} \\
\hline
\begin{minipage}[t]{0.92\textwidth}
\vspace{0.1em}
\begin{verbatim}
EXPERIMENT_PROMPT: 
  You are writing a Python script named `experiment.py` that must be runnable.

  ## Research Context
  Title: {title}
  Problem: {problem}
  Novelty: {novelty}
  Proposed Approach: {approach}

  ## Experimental Setup
  The following describes the experiment setup. You 
  must base your implementation strictly on this structure:

  Models/Algorithms to use: {model}
  Datasets involved: {dataset}
  Evaluation metrics: {metric}

  ## Execution Command (DO NOT MODIFY):
  You have {max_runs} runs to complete this 
  experiment. For each run, the script will be executed using:
  `python experiment.py --out_dir=run_i`
  where `i` is the run number (`run_1`, `run_2`, etc.).

  ## YOU MUST ENSURE experiment.py:
  1. Parses the `--out_dir` argument.
  2. Creates the output directory using `os.makedirs(out_dir, exist_ok=True)`.
  3. Performs actual model training and evaluation — 
  DO NOT simulate results using random numbers or 
  hardcode experiment result, all result should get from execution.
  4. Implements evaluation metircs with real logic.
  5. **Saves results as a dictionary in a file named 
  `final_info.json` placed directly inside 
  `out_dir`** — do **not** save into nested folders like `out_dir/variant_name/final_info.json`.

  ## Computational Constraints
  - Ensure the code is computationally affordable to run on a single GPU or CPU machine.
  - Avoid using large models like GPT, T5, BERT-large, or full ImageNet training.
  - Prefer small-scale tasks, toy models, or low-
  cost benchmarks (e.g., MNIST, UCI datasets, small MLPs or ResNet18).
  - Do not use complex distributed training or multi-GPU setups.

  Do not add extra command-line arguments.
  If your current experiment.py has placeholder code
  like `...`, replace them with runnable implementations.
  If any external functions like `compute_loss`, 
  `evaluate_model`, or `log_results` are used, implement them too.

  ## Baseline Results
  You do not need to re-run the baseline. 
  If available, the results are provided below:
  {baseline_results}

  ---
  Please begin writing the `experiment.py` file now.

\end{verbatim}
\vspace{0.1em}
\end{minipage} \\
\hline
\end{tabular}
\end{table}

\clearpage
\begin{table}
\centering
\caption{\textbf{Coder execution and error handling prompt}.}
\label{tab:error_handling_prompt}
\footnotesize
\begin{tabular}{|p{0.95\textwidth}|}
\hline
\textbf{Coder Execution and Error Handling Prompt} \\
\hline
\begin{minipage}[t]{0.92\textwidth}
\vspace{0.1em}
\begin{verbatim}
EXPERIMENT_SUCCESS_PROPMT: 
  Run {run_num} completed. Here are the results:
  {results}

  Decide if you need to re-plan your experiments given the result (you often will not need to).

  Someone else will be using `notes.txt` to perform a writeup on this in the future.
  Please include *all* relevant information for the 
  writeup on Run {run_num}, including an experiment 
  description and the run number. Be as verbose as necessary.

  Then, implement the next thing on your list.
  We will then run the command `python experiment.py --out_dir=run_{next_run}'.
  YOUR PROPOSED CHANGE MUST USE THIS COMMAND FORMAT, DO NOT ADD ADDITIONAL COMMAND LINE ARGS.
  If you are finished with experiments, respond with 'ALL_COMPLETED'.

EXPERIMENT_FAILURE_PROMPT: 
  There was an error running the experiment script:
  {message}
  Your goal is still to implement this experiment: {Title}.
  The purpose is: {Experiment}.
  You have {max_runs} runs total. We're currently on run {run_time}.
  Please fix `experiment.py` so that it runs successfully with:
  `python experiment.py --out_dir=run_{run_time}`.
  Make sure to implement any missing parts like 
  model definition, loss function, data loading, and final_info.json saving.

\end{verbatim}
\vspace{0.1em}
\end{minipage} \\
\hline
\end{tabular}
\end{table}

\clearpage
\begin{table}
\centering
\caption{\textbf{Coder experiment format prompt}.}
\label{tab:experiment_format_prompt}
\footnotesize
\begin{tabular}{|p{0.95\textwidth}|}
\hline
\textbf{Coder Experiment Format Prompt} \\
\hline
\begin{minipage}[t]{0.92\textwidth}
\vspace{0.1em}
\begin{verbatim} 
EXPERIMENT_FORMAT_PROMPT: 
  The experiment is organized into three sections:
  ## Model Section:
  {model}

  ## Dataset Section:
  {dataset}

  ## Metric Section:
  {metric}

  Your job is to extract the essential names of 
  models, datasets, and evaluation metrics that are directly useful for coding and experimentation.

  ### Output Format:
  Return a JSON object with the following structure:
  ```json
  {{
    "model": ["Model1", "Model2", ...],
    "dataset": ["Dataset1", "Dataset2", ...],
    "metric": ["Metric1", "Metric2", ...]
  }}

\end{verbatim}
\vspace{0.1em}
\end{minipage} \\
\hline
\end{tabular}
\end{table}

\clearpage
\begin{table}
\centering
\caption{\textbf{Writer system prompt}.}
\label{tab:writer_system_prompt}
\small 
\begin{tabular}{|p{0.95\textwidth}|}
\hline
\textbf{Writer System Prompt} \\
\hline
\begin{minipage}[t]{0.92\textwidth}
\vspace{0.05em}
\begin{verbatim}
WRITE_SYSTEM_PROMPT:
  You are an ambitious AI PhD student who is looking to publish a paper 
  that will contribute significantly to the field.
  You have already figured out the research idea and the experiments you 
  want to run.

  IMPORTANT - Citation Principles:
  - Each citation must support a specific claim, statement, or finding 
  in that sentence
  - Citations should be placed right after the claim they support
  - Do NOT add citations randomly or at the end of unrelated sentences
  - Only cite papers that are directly relevant to the specific content 
  of that sentence
  - Use \cite{{}} or \citet{{}} to reference papers in references.bib

  CRITICAL - LaTeX Format Only (NO Markdown):
  - NEVER mix Markdown and LaTeX syntax (e.g., **text\textbf{{ is WRONG)
  - Use \textbf{{text}} for bold, NEVER **text**
  - Use \textit{{text}} for italic, NEVER *text*
  - Use \texttt{{text}} for code/monospace, NEVER `text`
  - Example: \textbf{{Data Collection}}: not **Data Collection**: or **Data Collection\textbf{{
  - When including tables, always use proper LaTeX tabular format (not Markdown)
  - Avoid using Markdown-style tables (e.g., those starting with `| 
  Column |`) — they are not compatible with LaTeX rendering and will break the document
  - For algorithms, use \begin{{algorithmic}}...\end{{algorithmic}} with 
  \STATE, \FOR{{...}}, \IF{{...}}, \WHILE{{...}}, \ENDFOR, \ENDIF, 
  \ENDWHILE (uppercase commands)

  [LaTeX Formatting Reminder]
  - Use `\%` to indicate percentage values (e.g., 93\%)
  - Do not escape comment `%` symbols (e.g., `% comment`)
  - Wrap math expressions with `$...$`
  - Escape special characters, `_` as `\_`, `&` as `\&`, `#` as `\#`, etc.
  - For algorithms, use uppercase commands like:
    \begin{{algorithm}}
    \caption{{Algorithm Name}}
    \begin{{algorithmic}}
    \STATE Initialize parameters $\theta$
    \FOR{{each iteration $t = 1$ to $T$}}
        \STATE Update $\theta$ using gradient
    \ENDFOR
    \end{{algorithmic}}
    \end{{algorithm}}

  [Equation Length Guidelines - CRITICAL]
  - NEVER write equations that exceed one line width
  - For long equations, use multi-line environments:
     \begin{{align}}...\end{{align}} for multiple aligned equations
     \begin{{multline}}...\end{{multline}} for single equation split across lines
     \begin{{split}}...\end{{split}} inside equation environment
  - Break equations at operators (+, -, =, \cdot, etc.)
  - Use concise notation: prefer $h_t$ over $h_{{timestep}}$
  - Example of proper line breaking:
    \begin{{align}}
    \mathcal{{L}} &= \sum_{{i=1}}^N \left( f(x_i; \theta) - y_i \right)^2 \\
                  &\quad + \lambda \|\theta\|_2^2
    \end{{align}}
\end{verbatim}
\vspace{0.1em}
\end{minipage} \\
\hline
\end{tabular}
\end{table}
\vspace{-0.5em}

\clearpage
\begin{table}
\centering
\caption{\textbf{Writer section writing tips prompt}.}
\label{tab:section_tips_prompt}
\footnotesize
\begin{tabular}{|p{0.95\textwidth}|}
\hline
\textbf{Writer Section Writing Tips Prompt} \\
\hline
\begin{minipage}[t]{0.92\textwidth}
\vspace{0.1em}
\begin{verbatim}
SECTION_TIPS:
    Abstract:
    - TL;DR of the paper
    - What are we trying to do and why is it relevant?
    - Why is this hard?
    - How do we solve it (i.e. our contribution!)
    - How do we verify that we solved it (e.g. Experiments and results)
    
    Please make sure the abstract reads smoothly and is well-motivated. 
    This should be one continuous paragraph with no breaks between the lines.
    
    Introduction:
    - Write 5 paragraphs for the instructions. Each paragraph should answer one question.
    - paragraph1: What is the problem?
    - paragraph2: Why is it interesting and important?
    - paragraph3: Why is it hard? (E.g., why do naive approaches fail?)
    - paragraph4: Why hasn't it been solved before?
          (Or, what's wrong with previous proposed solutions? How does mine differ?)
    - paragrap5: What are the key components of my approach and results?
    
    Related_Work:
    - Academic siblings of our work, i.e. alternative attempts in literature 
      at trying to solve the same problem.
    - Goal is to "Compare and contrast" - how does their approach differ in 
      either assumptions or method?
    - Note: Just describing what another paper is doing is not enough. We need to compare and contrast.
    - STRUCTURE REQUIREMENT: Write 3 paragraphs (at most 4 paragraphs). Each 
    paragraph should focus on a distinct category or theme of related work.
    
    Method:
    - Start with Problem Definition (First Paragraph)
    - Use Math to Describe the Algorithm / Model
    - Explain the Motivation for Each Design Choice
    - Focus on the Core Algorithm / Architecture / Key Innovations
    - Avoid Experimental Details
    - End with a High-Level Summary (Optional)
    
    Experimental_Setup:
    - How do we test that our stuff works? Introduces a specific 
    instantiation of the Problem Setting and specific implementation details 
    of our Method for this Problem Setting.
    - Do not imagine unknown hardware details.
    - Includes a description of the dataset, evaluation metrics, important 
    hyperparameters, and implementation details.
    
    Results:
    - Shows the results of running Method on our problem described in 
    Experimental Setup.
    - Statement should be emphasized and as the paragraph title
    - You need to first describe the trends and improvement ratio and change 
    in the Table. include all the basic information
    - Furthermore, you need to analyze and provide reasons about why this 
    works or not
    
    Discussion:
    - Each subsection should be a question that challenges your method
    - Be specific and quantitative in your defense (cite figures, tables, numbers)
    
    Conclusion:
    - KEEP IT SHORT: ONE single paragraph only (~100-150 words)
    - Briefly restate the problem and key contribution (1-2 sentences)
    - Summarize main findings (1-2 sentences)
    - Mention one limitation or future direction (1 sentence)
    - Do NOT repeat detailed content from Discussion or Results
    - Academic papers favor concise, single-paragraph conclusions
\end{verbatim}
\vspace{0.1em}
\end{minipage} \\
\hline
\end{tabular}
\end{table}

\clearpage
\begin{table}
\centering
\caption{\textbf{Writer abstract writing prompt}.}
\label{tab:abstract_instruction_prompt}
\footnotesize
\begin{tabular}{|p{0.95\textwidth}|}
\hline
\textbf{Writer Abstract Writing Prompt} \\
\hline
\begin{minipage}[t]{0.92\textwidth}
\vspace{0.1em}
\begin{verbatim}
Abstract_Prompt: |
  You are writing the Abstract section of a top-tier AI research paper.
  Some tips are provided below:
  {abstract_tips}

  Here is the research idea that the paper is based on:
  {idea}

  IMPORTANT: The full paper has already been written. 
  Below is the complete content of all sections.
  Your task is to write an abstract that accurately summarizes 
  what was actually written in the paper,
  not just the initial idea. Pay attention to the actual methods used, 
  experiments conducted, and results obtained.

  === FULL PAPER CONTENT ===
  {full_paper_content}
  === END OF PAPER CONTENT ===

  Based on the full paper content above, write a concise abstract that:
  1. States the problem addressed in the paper
  2. Explains why it's important and challenging
  3. Describes the proposed solution/method
  4. Summarizes the key experiments and results
  5. Highlights the main contributions

  The output must be pure LaTeX and enclosed with \begin{{abstract}} ... \end{{abstract}}.
  Be sure to first name the file and use *SEARCH/REPLACE* blocks to perform these edits.

\end{verbatim}
\vspace{0.1em}
\end{minipage} \\
\hline
\end{tabular}
\end{table}

\begin{table}
\centering
\caption{\textbf{Writer citation system prompt}.}
\label{tab:citation_system_prompt}
\footnotesize
\begin{tabular}{|p{0.95\textwidth}|}
\hline
\textbf{Writer Citation System Prompt} \\
\hline
\begin{minipage}[t]{0.92\textwidth}
\vspace{0.1em}
\begin{verbatim}
CITATION_SYSTEM_PROMPT: |
  You are an academic writing assistant helping add and embed citation coverage in a research paper.

  Your role:
  - When asked to suggest citations, return only real, published academic 
  paper titles that are highly relevant to the given content.
  - When asked to embed citations, you will be provided with papers in the 
  format: [CITE_KEY: bibtex_key] Paper Title (authors, venue, year)
  - You MUST extract and use the EXACT bibtex_key from [CITE_KEY: 
  bibtex_key] when inserting `\cite{{bibtex_key}}` commands.
  - For example, if you see [CITE_KEY: smith2023nlp] Natural Language 
  Processing..., you must write \cite{{smith2023nlp}}
  - NEVER invent citation keys - only use the keys explicitly provided in 
  [CITE_KEY: ...]

  CRITICAL - MEANINGFUL CITATION PLACEMENT:
  - Each citation MUST support a specific claim, method, finding, or 
  statement in that sentence
  - Citations should be placed immediately after the claim they support
  - Do NOT randomly insert citations without clear relevance to the sentence 
  content
  - If a paper is not directly relevant to a specific claim, do NOT force it 
  into that sentence
  - Each citation should have a clear purpose: supporting a claim, 
  referencing a method, citing prior work, etc.

  Do not invent or fabricate any citations.
  Do not output BibTeX, author names, or publication details.

  CRITICAL - LaTeX Format Only (NO Markdown):
  - Use \textbf{{text}} for bold, NOT **text**
  - Use \textit{{text}} for italic, NOT *text*
  - Use \texttt{{text}} for code/monospace, NOT `text`
  Be thorough - missing citations is not acceptable
  Always follow the expected output format (JSON array or updated LaTeX 
  content), with no extra commentary or explanation.

\end{verbatim}
\vspace{0.1em}
\end{minipage} \\
\hline
\end{tabular}
\end{table}

\clearpage
\begin{table}
\centering
\caption{\textbf{Writer citation collection prompt}.}
\label{tab:citation_collection_prompt}
\footnotesize
\begin{tabular}{|p{0.95\textwidth}|}
\hline
\textbf{Writer Citation Collection Prompt} \\
\hline
\begin{minipage}[t]{0.92\textwidth}
\vspace{0.1em}
\begin{verbatim}
ADD_NEW_CITATIONS_PROMPT:
  You are adding NEW citations to the {section} section.

  Current section content:
  {section_content}

  NEW papers to cite (cite ALL of these):
  {paper_context}

  CRITICAL INSTRUCTIONS:
  1. Each paper is formatted as: [CITE_KEY: bibtex_key] Title (authors, venue, year)
  2. You MUST extract the bibtex_key from [CITE_KEY: bibtex_key] and use it for citations
  3. Add citations using \\cite{{bibtex_key}} format (use EXACTLY the bibtex_key shown)
  4. Cite ALL {num_papers} new papers provided above
  5. Keep ALL existing citations and content intact
  6. Add new citations in contextually appropriate places
  7. If needed, expand 1-2 sentences to naturally incorporate citations
  8. Do NOT remove or change existing citations
  9. Example: If you see [CITE_KEY: Smith2024], cite it as \\cite{{Smith2024}}

  MEANINGFUL CITATION PLACEMENT:
  - Each citation MUST directly support the specific claim or statement in that sentence
  - Place citations immediately after the claim they support (e.g., "Recent work shows 
  improvement~\\cite{{Smith2024}}.")
  - Read each paper's abstract to understand what it claims, then cite it  where that 
  claim is relevant
  - Do NOT add citations randomly or at the end of unrelated sentences
  - If you need to cite a paper but no suitable sentence exists, add a new  sentence that
  makes a claim the paper supports
  - Example: If a paper proposes a new architecture, cite it when discussing  that 
  architecture, NOT when discussing datasets

  Output the enhanced section with all citations.
\end{verbatim}
\vspace{0.1em}
\end{minipage} \\
\hline
\end{tabular}
\end{table}

\begin{table}[H]
\centering
\caption{\textbf{Writer citation search prompt}.}
\label{tab:citation_embedding_prompt}
\footnotesize
\begin{tabular}{|p{0.95\textwidth}|}
\hline
\textbf{Writer Citation Search Prompt} \\
\hline
\begin{minipage}[t]{0.92\textwidth}
\vspace{0.1em}
\begin{verbatim}
CITATION_SEARCH_QUERY_PROMPT:
  You will generate search queries to find relevant papers for citation in the {section} section.

  CONTEXT:
  - Idea (single string):
    {idea}
  - Section Content Snippet:
    {snippet}

  YOUR TASK: Read the IDEA and SNIPPET above and extract 2-3 SPECIFIC technical terms 
  FROM THEM to create 6-10 search queries.

  CRITICAL STEP-BY-STEP:
  1. READ the idea and the section snippet carefully
  2. IDENTIFY specific techniques/algorithms/models (e.g., "graph attention network", 
  "knowledge distillation", "BERT")
  3. IDENTIFY the task/application when present (e.g., "link prediction", "question 
  answering")
  4. CREATE queries as: "specific_technique AND specific_task" (if task is present), 
  otherwise combine 2-3 specific techniques
  5. DO NOT use generic terms; use only concrete technical terminology

  TYPES OF SPECIFIC TERMS TO EXTRACT:
  - Specific models: "BERT", "GPT-4", "T5", "ResNet"
  - Specific architectures: "graph attention network", "encoder-decoder"
  - Specific techniques: "knowledge distillation", "contrastive learning", "adversarial 
  training"
  - Specific mechanisms: "multi-head attention", "cross-attention", "tree-based reasoning"
  - Specific tasks: "question answering", "sentiment analysis", "link prediction"
  - Specific datasets/benchmarks: "SQuAD", "GLUE", "ImageNet"

  WHAT TO AVOID:
  - "deep learning", "machine learning", "AI"
  - "neural networks" (specify which type)
  - Domain-only terms: "NLP", "CV"
  - Vague phrases: "improve performance", "state-of-the-art"

  FORMAT EXAMPLES (structure only):
  - If idea/snippet mentions "BERT for question answering" → "BERT AND question answering"
  - If mentions "graph attention on citation networks" → "graph attention network AND 
  citation network"
  - If discusses "knowledge distillation for language models" → "knowledge distillation 
  AND language model"

  OUTPUT FORMAT:
  Return ONLY a JSON array of 6-10 query strings. No extra text, no markdown, no explanations.

  Example:
  [
    "graph attention network AND link prediction",
    "knowledge distillation AND language model",
    "multi-head attention AND machine translation"
  ]
\end{verbatim}
\vspace{0.1em}
\end{minipage} \\
\hline
\end{tabular}
\end{table}

\clearpage
\begin{table}[H]
\centering
\caption{\textbf{Writer refinement prompt}.}
\label{write-refinement}
\footnotesize
\begin{tabular}{|p{0.95\textwidth}|}
\hline
\textbf{Writer Refinement Prompt} \\
\hline
\begin{minipage}[t]{0.92\textwidth}
\vspace{0.1em}
\begin{verbatim}
MULTI_ROUND_REFINEMENT_PROMPT:
You are refining the {section} section (Round {round_num}/{total_rounds}).

  Current Focus for This Round: {focus}

  Current Section Content:
  {section_content}

  Section Guidelines:
  {section_tips}

  SPECIAL INSTRUCTION for Conclusion: If this is the Conclusion section, you MUST 
  keep it as ONE single paragraph (~100-150 words). Do NOT expand it. Do NOT add multiple paragraphs.

  Full Paper Context (all other sections for coherence check):
  {other_sections_context}

  Refinement Instructions:
  1. {method_specific_instruction}
  2. Improve mathematical notation and formalism
  3. Add design rationale for key choices
  4. CRITICAL: Read the Full Paper Context above to ensure your refined section is 
  coherent with other sections
  5. Ensure terminology, notation, and narrative flow are consistent across the paper

  CRITICAL - CITATION PRESERVATION:
  6. PRESERVE ALL EXISTING \\cite{{}} COMMANDS unless their supporting content is removed
  7. If you keep content, you MUST keep its citations
  8. If you remove redundant/contradictory content, you may remove its citations too
  9. Do NOT arbitrarily relocate citations to unrelated sentences
  10. If you add new content, add appropriate NEW citations to support it
  11. Aim to maintain or increase the total number of citations 
  (unless removing redundant sections)

  MEANINGFUL CITATION PLACEMENT:
  - Each citation must support a specific claim, statement, or finding in that sentence
  - Citations should be placed right after the claim they support
  - Do NOT add citations randomly or where they don't support the sentence content

  Common Errors to Avoid:
  {error_list}

  CRITICAL - LaTeX Format Only (NO Markdown):
  - Use \textbf{{text}} for bold, NOT **text**
  - Use \textit{{text}} for italic, NOT *text*
  - Use \texttt{{text}} for code/monospace, NOT `text`
  - For algorithms, use \STATE, \FOR{{...}}, \ENDFOR, \IF{{...}}, \ENDIF (uppercase)

  Output the refined section directly (no explanations).
\end{verbatim}
\vspace{0.1em}
\end{minipage} \\
\hline
\end{tabular}
\end{table}

\begin{table}[H]
\centering
\footnotesize
\caption{\textbf{Reviewer system prompt}.}
\label{tab:reviewer_system_prompt}
\begin{tabular}{|p{0.95\textwidth}|}
\hline
\textbf{Reviewer System Prompt} \\
\hline
\begin{minipage}[t]{0.92\textwidth}
\vspace{0.1em}
\begin{verbatim}
REVIEWER_SYSTEM_PROMPT_BASE: 
  You are an AI researcher who is reviewing a paper 
  that was submitted to a prestigious ML venue. Be critical and cautious in your decision.
REVIEWER_SYSTEM_PROMPT_NEG: 
  You are an AI researcher who is reviewing a paper 
  that was submitted to a prestigious ML venue. Be 
  critical and cautious in your decision. If a 
  paper is bad or you are unsure, give it bad scores and reject it.
REVIEWER_SYSTEM_PROMPT_POS: 
  You are an AI researcher who is reviewing a paper 
  that was submitted to a prestigious ML venue. Be 
  critical and cautious in your decision. If a 
  paper is good or you are unsure, give it good scores and accept it.
METAREVIEW_SYSTEM_PROMPT: 
  You are an Area Chair at a machine learning conference.
  You are in charge of meta-reviewing a paper that was reviewed by {reviewer_count} reviewers.
  Your job is to aggregate the reviews into a single meta-review in the same format.
  Be critical and cautious in your decision, find consensus, and respect the opinion of all the
  reviewers.

\end{verbatim}
\vspace{0.1em}
\end{minipage} \\
\hline
\end{tabular}
\end{table}

\clearpage
\begin{table}
\centering
\caption{\textbf{Reviewer format control prompt}.}
\label{tab:reviewer_template_prompt}
\footnotesize
\begin{tabular}{|p{0.95\textwidth}|}
\hline
\textbf{Reviewer Format Control Prompt} \\
\hline
\begin{minipage}[t]{0.92\textwidth}
\vspace{0.1em}
\begin{verbatim}
TEMPLATE_INSTRUCTIONS:
  Respond in the following format:

  THOUGHT:
  <THOUGHT>

  REVIEW JSON:
  ```json
  <JSON>
  ```

  In <THOUGHT>, first briefly discuss your intuitions and reasoning for the evaluation.
  Detail your high-level arguments, necessary choices and desired outcomes of the review.

  Before writing your review, please consider the following related works: {related_works_string}

  Do not make generic comments here, but be specific to your current paper.
  Treat this as the note-taking phase of your review.

  In <JSON>, provide the review in JSON format with the following fields in the order:
  - "Summary": A summary of the paper content and its contributions.
  - "Strengths": A list of strengths of the paper.
  - "Weaknesses": A list of weaknesses of the paper.
  - "Originality": A rating from 1 to 4 (low, medium, high, very high).
  - "Quality": A rating from 1 to 4 (low, medium, high, very high).
  - "Clarity": A rating from 1 to 4 (low, medium, high, very high).
  - "Significance": A rating from 1 to 4 (low, medium, high, very high).
  - "Questions": A set of clarifying questions to be answered by the paper authors.
  - "Limitations": A set of limitations and potential negative societal impacts of the work.
  - "Ethical Concerns": A boolean value indicating whether there are ethical concerns.
  - "Soundness": A rating from 1 to 4 (poor, fair, good, excellent).
  - "Presentation": A rating from 1 to 4 (poor, fair, good, excellent).
  - "Contribution": A rating from 1 to 4 (poor, fair, good, excellent).
  - "Overall": A rating from 1 to 10 (very strong reject to award quality).
  - "Confidence": A rating from 1 to 5 (low, medium, high, very high, absolute).
  - "Decision": A decision that has to be one of the following: Accept, Reject.

\end{verbatim}
\vspace{0.1em}
\end{minipage} \\
\hline
\end{tabular}
\end{table}

\clearpage
\begin{table}
\centering
\caption{\textbf{Reviewer reflection prompt}.}
\label{tab:reviewer_reflection_prompt}
\footnotesize
\begin{tabular}{|p{0.95\textwidth}|}
\hline
\textbf{Reviewer Reflection Prompt} \\
\hline
\begin{minipage}[t]{0.92\textwidth}
\vspace{0.1em}
\begin{verbatim}
REVIEW_REFLECTION_PROMPT: 
  In your thoughts, first carefully consider the accuracy and soundness of the review you just created.
  Include any other factors that you think are important in evaluating the paper.
  Ensure the review is clear and concise, and the JSON is in the correct format.
  Do not make things overly complicated.
  In the next attempt, try and refine and improve your review.
  Stick to the spirit of the original review unless there are glaring issues.

  Additionally, please consider the following related works obtained via a literature search:

  ```
  {related_works_string}
  ```

  Use these search results to assess the paper’s novelty, relevance, and significance.
  Provide specific comments on how the paper aligns with or differs from these related works.

  Respond in the same format as before:
  THOUGHT:
  <THOUGHT>

  REVIEW JSON:
  ```json
  <JSON>
  ```

  If there is nothing to improve, simply repeat the 
  previous JSON EXACTLY after the thought and 
  include "I am done" at the end of the thoughts but before the JSON.
  ONLY INCLUDE "I am done" IF YOU ARE MAKING NO MORE CHANGES.

\end{verbatim}
\vspace{0.1em}
\end{minipage} \\
\hline
\end{tabular}
\end{table}

\clearpage
\begin{table}
\centering
\caption{\textbf{Writing quality evaluation rubrics}.}
\label{tab:writing_quality_evaluation_rubrics}
\footnotesize
\begin{tabular}{|p{0.95\textwidth}|}
\hline
\textbf{Writing quality evaluation rubrics} \\
\hline
\begin{minipage}[t]{0.92\textwidth}
\vspace{0.1em}
\begin{verbatim}
You are an expert AI research reviewer evaluating the WRITING QUALITY and PRESENTATION of
a research paper.
**Paper Content:**
{paper_text}
**Your Task - Evaluate Writing Quality:**
  Quality Rubric (1–5 scale) focusing on Content Richness, References, and Writing:
  Score 1 — Very Poor: Critical information missing or wrong; methodology clearly infeasible or 
  incoherent.
    Examples: Missing problem statement, no clear methodology, completely unrealistic approach, 
              no literature review, no references to recent work.
  Score 2 — Poor: Key sections under-specified and novelty minimal; noticeable feasibility 
  or consistency issues.
    Examples: Vague problem description, weak novelty claim, unclear methodology, limited scope, 
              few or outdated references, minimal technical depth.
  Score 2.5 — Below Average: Basic structure present but lacks depth and recent references.
    Examples: Standard approach with minimal innovation, basic methodology, some references but not 
    cutting-edge, 
              feasible but not compelling, limited technical sophistication.
  Score 3 — Average: All required fields present and executable, but largely routine with limited 
  depth or insight.
    Examples: Standard approach, basic methodology, some novelty but not compelling, feasible but not 
    innovative, 
              references to well-known but not recent work.
  Score 3.5 — Above Average: Well-structured with some depth and recent references.
    Examples: Clear problem and methodology, some novel insights, references to recent papers 
    (2020-2023), 
              good feasibility, moderate technical sophistication.
  Score 4 — Good: Well-structured, mostly complete, shows some innovation; only minor gaps or risks 
  remain.
    Examples: Clear problem and methodology, some novel insights, well-defined scope, good 
    feasibility, references to recent and relevant work, good technical depth.
  Score 4.5 — Very Good: Strong structure with significant innovation and recent references.
    Examples: Highly innovative approach, comprehensive methodology, strong novelty, clear impact 
    potential, 
              references to cutting-edge work (2022-2024), excellent technical depth.
  Score 5 — Excellent: Original, rigorous, and comprehensive; strong experimental design and clear 
  path to publication.
    Examples: Highly innovative approach, comprehensive methodology, strong novelty, clear 
    impact potential, 
              references to state-of-the-art work, exceptional technical sophistication.
  Key Evaluation Criteria for Writing Quality:
    1. Content Richness: Depth of problem analysis, methodology detail, technical sophistication
    2. Reference Quality: Use of recent, relevant, and cutting-edge references (2020-2024 preferred)
    3. Writing Clarity: Clear exposition, logical flow, readability
    4. Technical Depth: Thoroughness of technical description and analysis
    5. Completeness: All necessary sections present and well-developed
**Output Format:**
Please respond with ONLY a JSON object:
{{
  "score": <number between 1 and 5, can use 0.5 increments>,
  "reason": "<detailed explanation focusing on writing quality, content depth, and references>"
}}
\end{verbatim}
\vspace{0.1em}
\end{minipage} \\
\hline
\end{tabular}
\end{table}

\clearpage
\begin{table}
\centering
\caption{\textbf{Idea quality evaluation rubrics}.}
\label{tab:idea_quality_evaluation_rubrics}
\footnotesize
\begin{tabular}{|p{0.95\textwidth}|}
\hline
\textbf{Idea quality evaluation rubrics} \\
\hline
\begin{minipage}[t]{0.92\textwidth}
\vspace{0.1em}
\begin{verbatim}
You are an expert AI research reviewer evaluating the RESEARCH VALUE and CONTRIBUTION 
of a research paper.
**Paper Content:**
{paper_text}
**Your Task - Evaluate Research Value:**
  Value Rubric (1–5 scale) focusing on Feasibility, Novelty, and Usefulness:
  Score 1 — Very Poor: Not feasible, no novelty, not useful.
    Examples: Unrealistic approach, duplicates existing work, provides no value to researchers.
  Score 2 — Poor: Major feasibility issues, minimal novelty, limited usefulness.
    Examples: Difficult to implement, minor variations of existing work, marginal contribution.
  Score 2.5 — Below Average: Some feasibility concerns, incremental novelty, modest usefulness.
    Examples: Technically possible but challenging, small improvements, some utility for specific 
            cases.
  Score 3 — Average: Feasible with effort, moderate novelty, decent usefulness.
    Examples: Standard approach with reasonable implementation, combines existing ideas in new ways, 
              useful for a subset of researchers.
  Score 3.5 — Above Average: Clearly feasible, good novelty, quite useful.
    Examples: Practical to implement, introduces interesting new perspectives, valuable for many 
              researchers in the field.
  Score 4 — Good: Highly feasible, significant novelty, very useful.
    Examples: Easy to implement and adopt, introduces novel techniques or insights, addresses 
              important problems, valuable for most researchers.
  Score 4.5 — Very Good: Excellent feasibility, strong novelty, highly useful.
    Examples: Ready-to-use solution, groundbreaking insights, solves critical problems, 
              game-changing for the field.
  Score 5 — Excellent: Perfect feasibility, exceptional novelty, transformative usefulness.
    Examples: Plug-and-play solution, paradigm-shifting ideas, revolutionizes the field, 
              essential for all researchers.
  Key Evaluation Criteria for Research Value:
    1. Feasibility: How practical and implementable is this research?
    2. Novelty: How original and innovative is the contribution?
    3. Usefulness: How valuable is this for researchers? Does it solve real problems?
    4. Impact Potential: Will this influence future research?
    5. Reproducibility: Can others easily build upon this work?
**Output Format:**
Please respond with ONLY a JSON object:
{{
  "score": <number between 1 and 5, can use 0.5 increments>,
  "reason": "<detailed explanation focusing on feasibility, novelty, and usefulness for researchers>"
}}
\end{verbatim}
\vspace{0.1em}
\end{minipage} \\
\hline
\end{tabular}
\end{table}

\clearpage
\begin{table}
\centering
\caption{\textbf{Case study on checker.} The checker is able to provide risk analysis of the task.}
\label{tab:safechecker_result}
\small
\begin{tabular}{|p{0.95\textwidth}|}
\hline
\textbf{SafeChecker Result} \\
\hline
\begin{minipage}[t]{0.92\textwidth}
\vspace{0.1em}
\begin{verbatim}
TASK_PROMPT: 
  Protecting Patient Data in Health Informatics Using Privacy-Preserving Technologies and 
  Secure Architecture
SafeChecker Result: 
  RISK_LEVEL: SAFE
  REASON: The prompt focuses on protecting patient data within health informatics systems 
  using privacy-preserving technologies and secure architectural designs. It promotes ethical 
  and responsible research aligned with improving cybersecurity, patient confidentiality, and 
  healthcare system integrity. There are no references to malicious techniques, bypassing 
  security systems, or any form of exploit, and it poses no risk of dual-use or harmful 
  application. This is a constructive and valuable research direction, especially relevant 
  to public health and information security.
  IS_ATTACKED: false
  ATTACK_TYPE: None
  REASON: The prompt "Protecting Patient Data in Health Informatics Using Privacy-Preserving 
  Technologies and Secure Architecture" is a benign, academic-style prompt that does not 
  contain any indicators of prompt attacks based on the listed categories.
\end{verbatim}
\vspace{0.1em}
\end{minipage} \\
\hline
\end{tabular}
\end{table}

\clearpage
\begin{table}
\centering
\caption{\textbf{Drawer generation prompt}.}
\label{tab:drawio_code_generator_prompt}
\footnotesize
\begin{tabular}{|p{0.95\textwidth}|}
\hline
\textbf{Drawer Generation Prompt} \\
\hline
\begin{minipage}[t]{0.92\textwidth}
\vspace{0.1em}
\begin{verbatim}
DRAWIO_CODE_GENERATOR_PROMPT:
You are a DrawIO code generator. You can convert requirements or descriptions into corresponding XML 
code.

## Core Capabilities
1. Generate runnable draw.io code directly from visual descriptions/requirements
2. Validation mechanism ensures code accuracy
3. Standardized code block output
4. Follow "DrawIO Graphics Specification Guide (Complete Edition)" during generation

## Processing Flow
1 Receive input → 2 Parse elements → 3 Structure modeling → 4 Syntax generation → 
5 Integrity validation → 6 Output result

## Output Specification
```xml
<!-- Validated draw.io code -->
<mxfile>
    [Generated core code]
</mxfile>
```
CRITICAL ID REQUIREMENTS:
Every mxCell element MUST have a unique id attribute
IDs must be alphanumeric and start with a letter
No empty or duplicate IDs allowed
Use descriptive IDs like "start_node", "process_step", "decision_point"
Root cells should have IDs "0" and "1"
All vertices and edges must have unique IDs
Interaction Rules
When receiving image descriptions: "Parsing structural relationships
(describing image details)...(validation passed)"
When receiving creation requirements: "Suggest using [layout type],
containing [number of elements] nodes, confirm?"
Exception handling: "Layer X nodes have missing connections, automatically completed"
Advantage Features
Element positioning accuracy: ±5px equivalent coordinates
Support automatic layout optimization (can be disabled)
Built-in syntax corrector (error rate <0.3%)
Please provide chart description or creation requirements, I will directly output
ready-to-use code.

Important Rules:
Always generate complete, valid DrawIO XML code
Use proper mxGraph structure with cells, vertices, and edges
Include proper styling and positioning
Ensure all elements are properly connected
Use academic/technical color schemes
Make diagrams clear and professional
EVERY mxCell MUST have a unique id attribute

\end{verbatim}
\vspace{0.1em}
\end{minipage} \\
\hline
\end{tabular}
\end{table}

\clearpage

\end{document}